\documentclass[10pt,journal,compsoc]{IEEEtran}

\ifCLASSOPTIONcompsoc
  \usepackage[nocompress]{cite}
\else
  \usepackage{cite}
\fi

\ifCLASSINFOpdf
\else
\fi

\usepackage{subfigure}
\usepackage{multirow}
\usepackage{graphicx}
\usepackage[space]{grffile}
\usepackage{mathtools, nccmath}
\usepackage{dblfloatfix}
\usepackage{amsmath}
\usepackage{hyperref}
\usepackage{dsfont}
\usepackage{wrapfig}
\usepackage{algorithm}
\usepackage{float}
\usepackage{algpseudocode}
\usepackage{algcompatible}
\usepackage{mathtools}
\usepackage{enumitem}
\usepackage{algorithmicx}
\usepackage{pifont}
\newcommand{\cmark}{\ding{51}}%
\newcommand{\xmark}{\ding{55}}%
\usepackage{xcolor}

\newcommand{\black}[1]{\textcolor{black}{#1}}

\begin{document}

\title{Dynamic Scheduling for Stochastic Edge-Cloud Computing Environments using A3C learning and Residual Recurrent Neural Networks}


\author{
        Shreshth~Tuli$^{*\dagger}$,
        Shashikant~Ilager$^*$,
        Kotagiri~Ramamohanarao$^*$,
    and~Rajkumar~Buyya$^*$
\IEEEcompsocitemizethanks{
\IEEEcompsocthanksitem All authors are with $^*$Cloud Computing and Distributed Systems (CLOUDS) Laboratory, School of Computing and Information Systems, The University of Melbourne, Australia\protect
\IEEEcompsocthanksitem S. Tuli is also with the $^\dagger$Department
of Computer Science and Engineering, Indian Institute of Technology, Delhi, India\protect\\
E-mail: shreshthtuli@gmail.com, shashikant.ilager@gmail.com, kotagiri@unimelb.edu.au and rbuyya@unimelb.edu.au}
\thanks{Manuscript received ---; revised ---.}}


\markboth{IEEE Transaction on Mobile Computing}%
{Tuli \MakeLowercase{\textit{et al.}}: Dynamic Scheduling for Stochastic Edge-Cloud Computing Environments}

\IEEEtitleabstractindextext{%
\begin{abstract}
The ubiquitous adoption of Internet-of-Things (IoT) based applications has resulted in the emergence of the Fog computing paradigm, which allows seamlessly harnessing both mobile-edge and cloud resources. Efficient scheduling of application tasks in such environments is challenging due to constrained resource capabilities, mobility factors in IoT, resource heterogeneity, network hierarchy, and stochastic behaviors. \black{Existing heuristics and Reinforcement Learning based approaches lack generalizability and quick adaptability, thus failing to tackle this problem optimally.  They are also unable to utilize the temporal workload patterns and are suitable only for centralized setups. However, Asynchronous-Advantage-Actor-Critic (A3C) learning is known to quickly adapt to dynamic scenarios with less data and Residual Recurrent Neural Network (R2N2) to quickly update model parameters. Thus, we propose an  A3C based real-time scheduler for stochastic Edge-Cloud environments allowing decentralized learning, concurrently across multiple agents. We use the R2N2 architecture to capture a large number of host and task parameters together with temporal patterns to provide efficient scheduling decisions.}  The proposed model is adaptive and able to tune different hyper-parameters based on the application requirements. We explicate our choice of hyper-parameters through sensitivity analysis. The experiments conducted on real-world data set show a significant improvement in terms of energy consumption, response time, Service-Level-Agreement and running cost by 14.4\%, 7.74\%, 31.9\%, and 4.64\%, respectively when compared to the state-of-the-art algorithms.
\end{abstract}

}

\maketitle

\IEEEdisplaynontitleabstractindextext

\IEEEpeerreviewmaketitle

\IEEEraisesectionheading{\section{Introduction}\label{sec:introduction}}
The advancements in the Internet of Things (IoT) have resulted in a massive amount of data being generated with enormous volume and rate. Applications that access this data, analyze and trigger actions based on stated goals, require adequate computational infrastructure to satisfy the requirements of users \cite{QoE}. Due to increased network latency, traditional cloud-centric IoT application deployments fail to provide quick response to many of the time-critical applications such as health-care, emergency response, and traffic surveillance \cite{Gubbi2013}. Consequently, emerging Edge-Cloud is a promising computing paradigm that provides a low latency response to this new class of IoT applications \cite{mcKinsey, TULI201922, wang2019delay}.  Here, along with remote cloud, the edge of the network have limited computational resources to provide a quick response to time-critical applications.     

The resources at the edge of the network are constrained due to cost and feasibility factors \cite{chen2015efficient}. Efficient utilization of Edge resources to accommodate a greater number of applications and to simultaneously maximize their Quality of Service (QoS) is extremely necessary. To achieve this, ideally, we need a scheduler that efficiently manages workloads and underlying resources. However, scheduling in the Edge computational paradigm is exceptionally challenging due to many factors. Primarily, due to the heterogeneity, computational servers between remote cloud and local edge nodes significantly differ in terms of their capacity, speed, response time, and energy consumption. Moreover, machines can also be heterogeneous within cloud and edge layers. Besides, due to the mobility factor in Edge paradigm, bandwidth continuously changes between the data source and computing nodes, which requires continual dynamic optimization to meet the application requirements. Furthermore, the Edge-Cloud environment is stochastic in many aspects, such as the task's arrival rate, duration of tasks, and their resource requirements, which further makes the scheduling problem challenging. Therefore, dynamic task scheduling to efficiently utilize the multi-layer resources in stochastic environments becomes crucial to save energy, cost and simultaneously improve the QoS of applications.   
  
The existing task or job scheduling algorithms in Edge-Cloud environments have been dominated by heuristics or rule-based policies \cite{skarlat2017optimized, beloglazov2012optimal, pham2017cost, brogi2017qos, choudhari2018prioritized, pham2016fog}. Although heuristics usually work well in general cases, they do not account for the dynamic contexts driven by both workloads and composite computational paradigms like Edge-Cloud.  Furthermore, they fail to adapt to continuous changes in the system \cite{JeffMLforSystem}, which is common in Edge-Cloud environments \cite{yi2015survey}. To that end, Reinforcement Learning (RL) based scheduling approach is a promising avenue for dynamic optimization of the system \cite{JeffMLforSystem, Fox2019 }.  The RL solutions are more accurate as the models are built from the actual measurements, and they can identify complex relationships between different interdependent parameters.  Recent works have explored different value-based RL techniques to optimize several aspects of Resource Management Systems (RMS) in distributed environments \cite{basu2019Megh, li2018IoTlearning,xu2017DLlaserdeadline, zhang2018doubleQLearningEdge }. Such methods store a Q value function in a table or using a Neural network for each state of the edge-cloud environment, which is an expected cumulative reward in the RL setup \cite{sutton1998introduction}. The tabular value-based RL methods face problem of limited scalability \cite{goodfellow2016deep, bowling2000convergence, van2016deep}, for which researchers have proposed various Deep learning based methods like Deep Q Learning (DQN) \cite{cheng2018drlcloud, mao2016RMSDRL, zhang2017energydeepQRL} which use a neural network to approximate the Q value. However, previous studies have shown that such value-based RL techniques are not suitable for highly stochastic environments \cite{mnih2016asynchronous}, which make them perform poorly in Edge-Cloud deployments. Limited number of works exist which are able to leverage policy gradient methods \cite{mao2016resource} and optimize for only a single QoS parameter and do not use asynchronous updates for faster adaptability in highly stochastic environments. Moreover, all prior works do not exploit temporal patterns in workload, network and node behaviours to further improve scheduling decisions. Furthermore, these works use a centralized scheduling policy which is not suitable for decentralized or hierarchical environments. Hence, this work maps and solves the scheduling problem in stochastic edge-cloud environments using asynchronous policy gradient methods which can recognize the temporal patterns using recurrent neural networks and continuously adapt to the dynamics of the system to yield better results.

In this regard, we propose a deep policy gradient based scheduling method to capture the complex dynamics of workloads and heterogeneity of resources. To continuously improve over the dynamic environment, we use the asynchronous policy gradient reinforcement learning method called Asynchronous Advantage Actor Critic (A3C). A3C, proposed by Mnih et al. \cite{mnih2016asynchronous}, is a policy gradient method for directly updating a stochastic policy which runs multiple actor-agents asynchronously with each agent having it's own neural network. The agents are trained in parallel and update a global network periodically, which holds shared parameters. After each update, the agents resets their parameters to those of the global network and continue their independent exploration and training until they update themselves again. \black{This method allows exploration of larger state-action space quickly~ \cite{mnih2016asynchronous} and enables models to rapidly adapt to stochastic environments. Moreover, it allows us to run multiple models asynchronously on different edge or cloud nodes in a decentralized fashion without a single point of failure.}\ Using this, we propose a learning model based on Residual Recurrent Neural Network (R2N2). \black{The R2N2 model is capable of accurately identifying the highly nonlinear patterns across different features of the input and exploiting the temporal workload and node patterns, with residual layers increasing the speed of learning \cite{yue2018residual}.}  Moreover, the \black{proposed scheduling model can be tuned to optimize the required QoS metrics based on the application demands using the adaptive loss function proposed in this work.}  To that end, minimizing this loss function through policy learning helps achieve highly optimized scheduling decisions. Unlike heuristics, the proposed framework can adapt to the new requirements as it continuously improves the model by tuning parameters based on new observations.  Furthermore, policy gradient enables our model to quickly adapt allocation policy responding to the dynamic workload, host behaviour and QoS requirements, compared to traditional DQN methods. The experiment results using an extended version of iFogSim Toolkit \cite{gupta2017ifogsim} with elements of CloudSim 5.0 \cite{calheiros2011cloudsim} show the superiority of our model against existing heuristics and previously proposed RL models. Our proposed methodology achieves significant efficiency for several critical metrics such as energy, response time, Service Level Agreements (SLA) violation \cite{beloglazov2012optimal} and cost among others.  

In summary, the \textbf{key contributions} of this paper are:      

  \begin{itemize}
\item We design an architectural system model for the data-driven deep reinforcement learning based scheduling for Edge-Cloud environments.   
\item We outline a generic \textit{asynchronous} learning model for scheduling in \textit{decentralized} environments.
\item We propose a \textit{Policy gradient} based Reinforcement learning method (A3C) for \textit{stochastic} dynamic scheduling method.
\item We demonstrate a \textit{Residual Recurrent Neural Network} (R2N2) based framework for exploiting temporal patterns for scheduling in a hybrid Edge-Cloud setup.
\item We show the superiority of the proposed solution through extensive simulation experiments and compare the results against several baseline policies.
\end{itemize}

The rest of the paper is organized as follows. Section \ref{sec:system_model} describes the system model and also formulates the problem specifications. Section \ref{sec:learning-model} explains a generic policy gradient based learning model. Section \ref{sec:stochastic-dynamic-scheduling} explains the proposed A3C-R2N2 model for scheduling in Edge-Cloud environments. The performance evaluation of the proposed method is shown in Section \ref{sec:perf-eval}. The relevant prior works are explained in Section \ref{sec:related-work}. Conclusions and future directions are presented in Section \ref{sec:conclusion}.

\begin{figure*}[t]
    \centering
    \includegraphics[width=1.7\columnwidth]{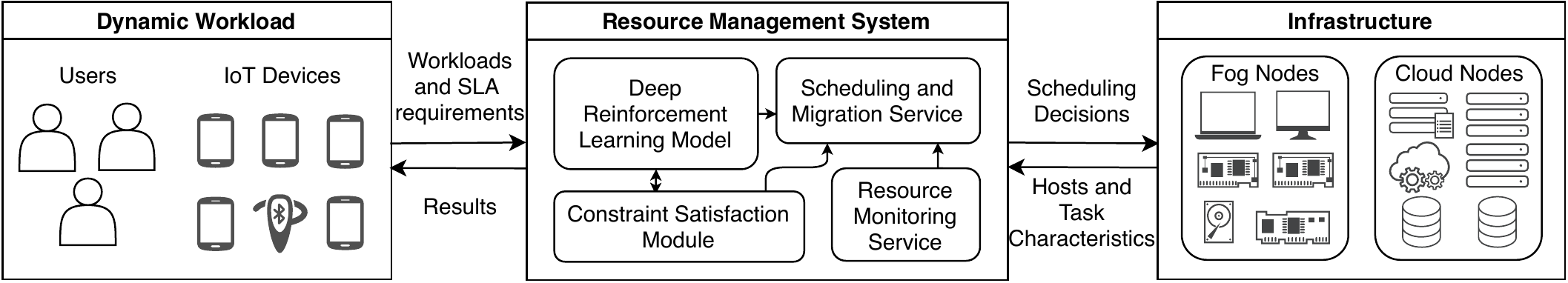}
    \caption{System Model}
    \label{fig:system}
\end{figure*}

\section{System Model and Problem Formulation}
\label{sec:system_model}

In this section, we describe the system model and interaction between various components that allow an adaptive reinforcement-based scheduling. In addition, we describe the workload model and problem formulation.

\subsection{System Model} 
\label{sec:system}

In this work, we assume that the underlying infrastructure is composed of both edge and cloud nodes. An overview of the system model is shown in Figure \ref{fig:system}. The edge-cloud environment consists of distributed heterogeneous resources in the network hierarchy, from the edge of the network to the multi-hop remote cloud. The computing resources act as hosts for various application tasks. These hosts can vary significantly in their compute power and response times. The edge devices are closer to the users and hence provide much lower response times but are resource-constrained with limited computation capability. On the other hand, cloud resources (Virtual Machines) located several hops away from the users, provide much higher response time. However, cloud nodes are resource enriched with increased computational capabilities that can process multiple tasks concurrently. 

The infrastructure is controlled by a Resource Management System (RMS) which consists of Scheduling, Migration and Resource Monitoring Services. The RMS receives tasks with their QoS and SLA requirements from IoT devices and users. It schedules the new tasks and also periodically decides if existing tasks needs to be migrated to new hosts based on the optimization objectives. The tasks' CPU, RAM, bandwidth, and disk requirements with their expected completion times or deadlines affect the decision of the RMS. This effect is simulated using a stochastic task generator known as the Workload Generation Module (WGM) following a dynamic workload model for task execution described in the next subsection. 

In our model, the Scheduler and Migration services interact with a Deep Reinforcement Learning Module (DRLM), which suggests placement decision for each task (on hosts) to the former services. Instead of a single scheduler, we run multiple schedulers with separate partitions of tasks and nodes. These schedulers can be run on a single node or separate edge-cloud nodes \cite{mnih2016asynchronous}. As shown in prior works \cite{mnih2016asynchronous, qi2018vehicular}, having multiple actors learn parameter updates in an asynchronous fashion allows computational load to be distributed among different hosts, allowing faster learning within the limits of resource constrained edge devices. Thus, in our system, we assume all edge and cloud nodes to accumulate local gradients to their schedulers and add and synchronize gradients of all such hosts to update their models individually. Our policy learning model is part of the DRLM with each scheduler with a separate copy of the global neural network, which allows asynchronous updates. Another vital component of the RMS is the Constraint Satisfaction Module (CSM) which checks if the suggestion from the DRLM is valid in terms of constraints such as whether a task is already in migration or the target host is running at full capacity. The importance and detailed functionality of CSM  is explained in Section \ref{sec:output}.

\subsection{Workload Model}
\label{sec:workload}

\begin{figure}[!b]
    \centering
    \includegraphics[width=0.9\columnwidth]{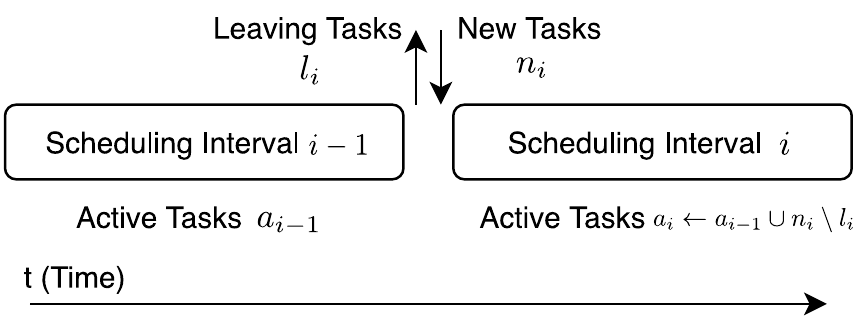}
    \caption{Dynamic Task Workload Model}
    \label{fig:workload}
\end{figure}

As described before, task generation is stochastic and each task has a dynamic workload. Based on changing user demands and mobility of IoT devices, the computation and bandwidth requirements of the tasks change with time. As done in prior works \cite{beloglazov2012optimal, gupta2017ifogsim}, we divide our execution time into scheduling intervals of equal duration. The scheduling intervals are numbered based on their order of occurrence as shown in Figure \ref{fig:workload}. The $i^{th}$ scheduling interval is shown as $SI_i$, which starts at time $t_i$ and continues till  the beginning of the next interval i.e., $t_{i+1}$. In each $SI_i$, the active tasks are those that were being executed on the hosts and are denoted as $a_i$. Also, at the beginning of $SI_i$, the set of tasks that get completed is denoted as $l_i$ and the new tasks that are sent by the WGM are denoted as $n_i$. The tasks $l_i$ leave the system and new tasks $n_i$ are added to the system. Thus, at the beginning of the interval $SI_i$, the active tasks $a_i$ is $a_{i-1} \cup n_i \setminus l_i$.

\subsection{Problem Formulation}
\label{sec:problem}

The problem that we consider is to optimize the performance of the scheduler in the edge-cloud environment as described in Section \ref{sec:system} and dynamic workload described in Section \ref{sec:workload}. The performance of the scheduler is quantified by the metric denoted as $Loss$ defined for each scheduling interval. The lower the value of $Loss$, the better the scheduler. We denote loss of the interval $SI_i$ as $Loss_i$.

In the  edge-cloud environment, the set of hosts is denoted as $Hosts$ and its enumeration as $[H_0,H_1,...,H_n]$. We assume that the maximum number of hosts at any instant of the execution is $n$. We also denote host assigned to a task $T$ as $\{T\}$. We define our scheduler as a mapping between the state of the system to an action which consists of host allocation for new tasks and migration decision for active tasks. The state of the system at the beginning of $SI_i$, denoted as $State_i$, consists of the parameter values of Hosts, remaining active tasks of the previous interval which ($a_{i-1}\setminus l_i$) and new tasks ($n_i$). The scheduler has to decide for each task in $a_i$ ($= a_{i-1} \cup n_i \setminus l_i$), the host to be allocated or migrated to, which we denote as the $Action_i$ for $SI_i$. However, all tasks may not be migratable. Let $m_i\subseteq a_{i-1}\setminus l_i$ be the migratable tasks. Thus, $Action_i = \{h\in Hosts \text{ for task }T\ |T\in m_i\cup n_i \}$ which is a migration decision for tasks in $m_i$ and allocation decision for tasks in $n_i$. Thus scheduler, denotes as $Model$, is a function$:State_i \rightarrow Action_i$. The $Loss_i$ of an interval depends on the allocation of the tasks to hosts i.e., $Action_i$ by the $Model$. Hence, for an optimal $Model$, the problem can be formulated as described by Equation \ref{eq:problem},
\begin{equation}
\label{eq:problem}
\begin{aligned}
& \underset{Model}{\text{minimize}}
& & \sum_i Loss_i \\
& \text{subject to}
& & \forall\ i,\ Action_i = Model(State_i)\\
&&& \forall\ i\ \forall\ T \in m_i \cup n_i, \{T\} \leftarrow Action_i(T).
\end{aligned}
\end{equation}

\begin{table}[]
    \centering
    \begin{tabular}{|c|c|}
        \hline
        Symbol & Meaning \\
        \hline \hline
        $SI_i$ & $i^{th}$ scheduling interval \\ \hline
        $a_i$ & Active tasks in $SI_i$ \\ \hline
        $l_i$ & Tasks leaving at beginning of $SI_i$ \\ \hline
        $n_i$ & New tasks received at beginning of $SI_i$ \\ \hline
        $Hosts$ & Set of hosts in the Edge-Cloud Datacenter \\ \hline
        $n$ & Number of hosts in the Edge-Cloud Datacenter \\ \hline
        $H_i$ & $i^{th}$ host in an enumeration of $Hosts$ \\ \hline
        $T_i^S$ & $i^{th}$ task in an enumeration of $S$ \\ \hline
        $\{T\}$ & Host assigned to task $T$ \\ \hline
        $FV_i^S$ & Feature vector corresponding to $S$ at $SI_i$\\ \hline
        $m_i$ & Migratable tasks in $a_i$\\ \hline
        $Action_i^{PG}$ & Scheduling decision at start of $SI_i$\\ \hline
        $Loss_i^{PG}$ & Loss function for the model at start of $SI_i$\\ \hline
    \end{tabular}
    \caption{Symbol Table}
    \label{tab:symbols}
\end{table}

\begin{figure}
    \centering
    \includegraphics[width=0.5\columnwidth]{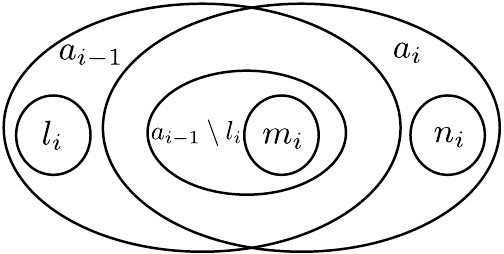}
    \caption{Venn Diagram of Various Task Sets}
    \label{fig:venn}
\end{figure}

A symbol table for ease of meaning recall and a Venn diagram of various task sets are given in Table \ref{tab:symbols} and Figure \ref{fig:venn}, respectively.

\section{Reinforcement Learning Model}
\label{sec:learning-model}

We now propose a Reinforcement Learning model for the problem statement described in Section \ref{sec:problem} suitable for policy gradient learning. First, we present the input and output specifications of the Neural Network and then describe the modeling of $Loss_i$ (from Equation \ref{eq:problem}) in our model.

\subsection{Input Specification}
\label{sec:input}

\begin{figure}[!b]
    \centering
    \subfigure[$FV_i^{Hosts}$]{
    \includegraphics[width=.27\columnwidth]{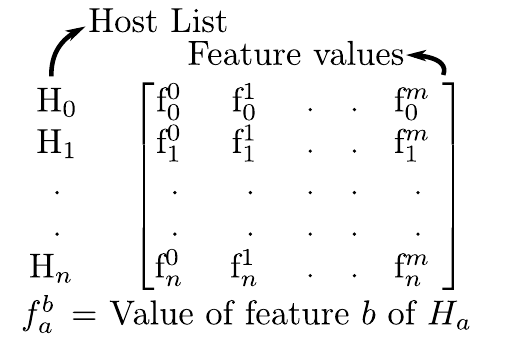}
    \label{fig:host-matrix}
    }
    \subfigure[$FV_i^{n_i}$]{
    \includegraphics[width=.27\columnwidth]{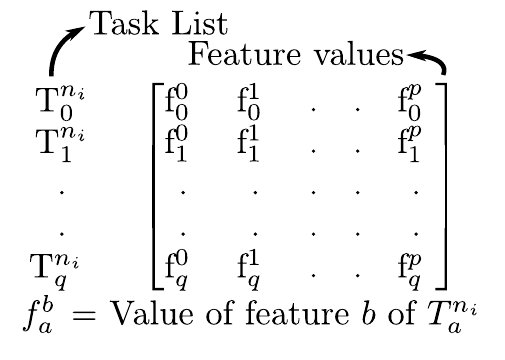}
    \label{fig:new-vm-matrix}
    }
    \subfigure[$FV_i^{a_{i-1}\setminus l_i}$]{
    \includegraphics[width=.36\columnwidth]{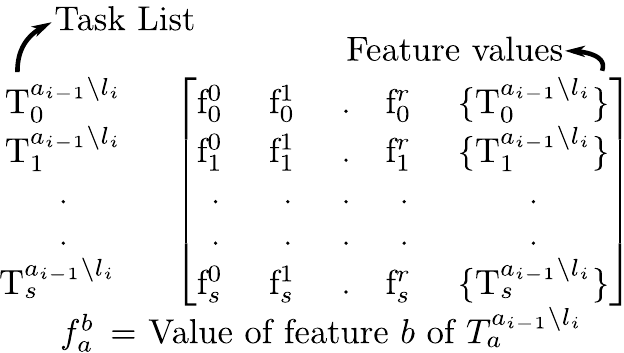}
    \label{fig:active-vm-matrix}
    }
    \caption{Matrix Representation of Model Inputs}
    \label{fig:input}
\end{figure}

The input of the scheduler $Model$, is the $State_i$ which consists of the parameters of hosts, which include utilization and capacity of CPU, RAM, bandwidth, and disk~\cite{basu2019Megh}. It also includes the power characteristics, cost per unit time, Million Instructions per Seconds (MIPS) for the host, response time, and the number of tasks to which this host is allocated. Different hosts would have different computational power (CPU), memory capacity (RAM) and I/O availability (disk and bandwidth). As tasks in an edge-cloud setup impose compute, memory and I/O limitations, such parameters are crucial for scheduling decisions. Moreover, allowing multiple tasks to be placed on a small cluster of hosts could ensure low energy usage (hibernating the ones with no tasks). A host with higher I/O capacity (disk read/write speeds) could allow I/O intensive tasks to be completed quickly and prevent SLA violations. All these parameters are defined for all hosts in a feature vector denoted as $FV_i^{Hosts}$ as shown in Figure \ref{fig:host-matrix}. The tasks in $a_i$ are segregated into two disjoint sets: $n_i$ and $a_{i-1}\setminus l_i$. The former consists of parameters like task CPU, RAM, bandwidth, and disk requirements. The latter also consists of the index of the host assigned in the previous interval. The feature vectors of these set of tasks are denoted as $FV_i^{n_i}$ and $FV_i^{a_{i-1}\setminus l_i}$ as shown in Figures \ref{fig:new-vm-matrix} and \ref{fig:active-vm-matrix} respectively. Thus, $State_i$ becomes $(FV_i^{Hosts},\,FV_i^{a_{i-1}\setminus l_i},\,FV_i^{n_i})$, which is the input of the model.

\subsection{Output Specification}

\label{sec:output}

At the beginning of the interval $SI_i$, the model needs to provide a host assignment for each task in $a_i$ based on the input $State_i$. The output, also denoted as $Action_i$ is a host assignment for each new task $\in n_i$ and migration decision for remaining active tasks from previous interval $\in a_{i-1} \setminus l_i$. This assignment must be valid in terms of the feasibility constraints such that each task which is migrated must be migratable to the new host (we denote migratable task as $m_i$ which is $\subseteq a_i$), i.e., it is not under migration. Moreover, when a host $h$ is allocated to any task $T$, then after allocation $h$ should not get overloaded i.e., $h$ is suitable for $T$. Thus, we describe $Action_i$ through Equation \ref{eq:action} such that for the interval $SI_i$, $\forall\ T \in n_i\cup m_i, \{T\} \leftarrow Action_i(T)$,

\begin{equation}
\label{eq:action}
\begin{aligned}
& Action_i = 
\begin{cases}
h \in Hosts\ \forall\ t \in n_i\\
h_{new} \in Hosts\ \forall\ t \in m_i\ if\ t\ is\ to\ be\ migrated\\
\end{cases} \\
& \text{subject to } \\
& \ \ \ Action_i \text{ is suitable for } t\ \forall\ t \in n_i \cup m_i.
\end{aligned}
\end{equation}

\begin{figure}[!b]
    \centering
    \includegraphics[width=0.45\columnwidth]{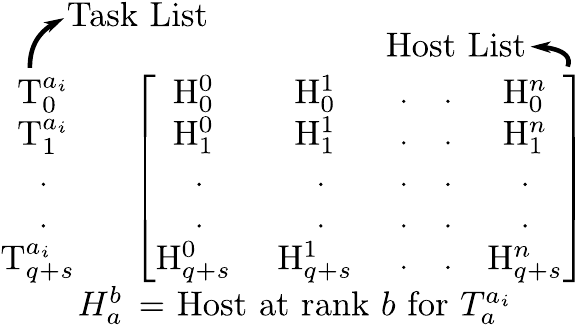}
    \caption{Matrix Representation of Model Output: $Action_i^{PG}$}
    \label{fig:action}
\end{figure}

However, developing a model that provides a constrained output is computationally difficult \cite{Pathak:2015:CCN:2919332.2919851} hence, we use an alternative definition of model action which is unconstrained. We compensate for the constraints in the objective function. In the unconstrained formulation of the model action, the output would be a priority list of hosts for each task. Thus, for task $T_j^{a_i}$, we have a list of hosts $[H_j^0, H_j^1, ..., H_j^n]$ in decreasing order of allocation preference. For a neural network, the output could be a vector of allocation preference for each host for every task. This means that rather than specifying a single host for each task, the model provides a ranked list of hosts. We denote this unconstrained model action for policy gradient setup as $Action_i^{PG}$ as shown in Figure \ref{fig:action}.

This unconstrained action cannot be used directly for updating the task allocation to hosts. We need to select the most preferable host for each task which is suitable for only those tasks that are migratable. To convert $Action_i^{PG}$ to $Action_i$ is straightforward as shown in Equation \ref{eq:action-selection}. For $Action_i(T_j^{a_i})$, if $T_j^{a_i} \in a_{i-1}\setminus l_i$ and is not migratable then it is not migrated. Otherwise, $T_j^{a_i}$ will be  allocated to the highest rank host which is suitable. By the conversion of Equation \ref{eq:action-selection}, $Action_i$ always obeys constraints specified in Equation \ref{eq:action} and hence is used for model update as

\begin{equation}
\label{eq:action-selection}
\begin{aligned}
Action_i(T_j^{a_i}) &= H_j^k |\  T_j^{a_i} \in m_i \cup n_i\\
& \wedge H_j^k \text{ is suitable for } T_j^{a_i}\\
& \wedge \forall\ l<k, H_j^l \in Action_{i-1}^{PG}(T_j^{a_i}), \\
& H_j^l \text{ is not suitable for } T_j^{a_i}.
\end{aligned}
\end{equation}

Additionally, we define penalty for the unconstrained action as in Equation \ref{eq:penalty}. This captures two aspects of penalty: (1) the \textit{migration penalty} as the fraction of tasks that the model wanted to migrate but cannot be migrated to the total number of tasks and (2) the \textit{host allocation penalty} as the sum for each task, the number of hosts that could not be allocated to that task but were given higher preference. This penalty would be used in the Loss function defined in Section \ref{sec:loss}. The first addend in Equation \ref{eq:penalty} captures the \textit{host allocation penalty} and the second addend captures the \textit{migration penalty} and this penalty guides the learning model to make decisions based on the constraints in Equation \ref{eq:action}. Thus, we define penalty as:

\begin{equation}
\label{eq:penalty}
\begin{aligned}
& Penalty_{i+1} = \\
& \frac{\sum_{t\in a_i} k\, | H^k = Action_i(t) \wedge H^k \in Action_i^{PG}(t)}{|a_i| \times n}\\
&+ \frac{\sum_{t\in a_{i-1}\setminus l_i}\mathds{1}(t \notin m_i \wedge Action_i(t) \neq \{t\} )}{|a_i|}.
\end{aligned}
\end{equation}

Hence, the output $Action_i^{PG}$ is first processed by the CSM to generate $Action_i$ and $Penalty_{i+1}$. Now, to update the parameters of the model at the beginning of $SI_i$, we incorporate both $Loss_i$ and $Penalty_i$ as described in the next subsection.

\subsection{Loss Function}
\label{sec:loss}

In our learning model, we want the model to be optimum to reduce $Loss_i$ in each interval and hence the cumulative loss. Also, we want our model, which is a mapping from $State_i$ to $Action_i$, to adapt to the dynamically changing state. For this, we now define $Loss_i$, which acts as a metric for parameter update for the model. First, we define various metrics (normalized to [0,1]) which help us to define $Loss_i$.

\begin{enumerate}[leftmargin=*]
    \item \textit{Average Energy Consumption} (AEC) is defined for any interval as the energy consumption of the infrastructure (which includes all edge and cloud hosts) normalized by the maximum power of the environment. However, edge and cloud nodes may have different energy sources like energy harvesting devices for edge and main supply for cloud \cite{roselli2015review}. Thus, we multiply the energy consumed by a host $h \in Hosts$ by a factor $\alpha_h \in [0,1]$ which can be set for edge and cloud nodes as per the user requirement and deployment strategy. The power is normalized as
    \begin{equation}
    \label{eq:aec}
        \begin{aligned}
        AEC_i^{Hosts} = \frac{\sum_{h\in Hosts} \alpha_h \int_{t = t_{i}}^{t_{i+1}} P_h(t)dt}{\sum_{h\in Hosts} \alpha_h  P_h^{max}(t_{i+1} - t_i)},
        \end{aligned}
    \end{equation}
    where $P_h(t)$ is the power function of host $h$ with time, and $P_h^{max}$ is maximum possible power of $h$.
    \item \textit{Average Response Time} (ART) is defined for an interval $SI_i$ as the average response time for all leaving tasks ($l_{i+1}$) in that interval normalized by maximum response time until  the current interval as shown in Equation \ref{eq:art}. The task response time is the sum of host (on which this task is scheduled) response time and task execution time. Hence ART is defined as
    \begin{equation}
    \label{eq:art}
        \begin{aligned}
        ART_i = \frac{\sum_{t \in l_{i+1}} Response\ Time(t)}{|l_{i+1}| \max_i \max_{t\in l_i} Response\ Time(t)}.
        \end{aligned}
    \end{equation}
    \item \textit{Average Migration Time} (AMT) is defined for an interval $SI_i$ as the average migration time for all active tasks ($a_i$) in that interval normalized by maximum migration time until the current interval as shown in Equation \ref{eq:amt}. AMT is defines as:
    \begin{equation}
    \label{eq:amt}
        \begin{aligned}
        AMT_i =  \frac{\sum_{t \in a_{i}} Migration\ Time(t)}{|a_{i}| \max_i \max_{t\in l_i} Response\ Time(t)}.
        \end{aligned}
    \end{equation}
    \item \textit{Cost} (C) is defined for an interval $SI_i$ as the total cost incurred during that interval as shown in Equation \ref{eq:cost}, 
    \begin{equation}
    \label{eq:cost}
        \begin{aligned}
        Cost_i =  \frac{\sum_{h\in Hosts} \int_{t = t_{i}}^{t_{i+1}} C_h(t)dt}{\sum_{h\in Hosts}  C_h^{max}(t_{i+1} - t_i)}.
        \end{aligned}
    \end{equation}
    where $C_h(t)$ is the cost function for host $h$ with time, and $C_h^{max}$ is maximum cost per unit for host $h$.
    \item \textit{Average SLA Violations} (SLAV) is defined for an interval $SI_i$  as the average number of SLA violations in that interval for leaving task ($l_{i+1}$) as shown in Equation \ref{eq:sla}. $SLA(t)$ of task $T$ is defined in \cite{beloglazov2012optimal} which is product of two metrics: (i) SLA violation time per active host and (ii) performance degradation due to migrations. Thus, 
    \begin{equation}
    \label{eq:sla}
        \begin{aligned}
        SLAV_i =  \frac{\sum_{t \in l_{i+1}} SLA(t)}{|l_{i+1}|}.
        \end{aligned}
    \end{equation}
\end{enumerate}

To minimize the above mentioned metrics, as done in various prior works~\cite{basu2019Megh, tuli2020healthfog}, we define $Loss_i$ as a convex combination of these metrics for interval $SI_{i-1}$. Thus,

\begin{equation}
\label{eq:loss}
\begin{aligned}
Loss_i &= \alpha \cdot AEC_{i-1} + \beta \cdot ART_{i-1} + \gamma \cdot AMT_{i-1} \\&+ \delta \cdot Cost_{i-1} + \epsilon \cdot SLAV_{i-1}\\
& \text{such that }\alpha, \beta, \gamma, \delta, \epsilon \geq 0\\
& \wedge \alpha+\beta+\gamma+\delta+\epsilon = 1.
\end{aligned}
\end{equation}

Based on different user QoS requirements and application settings different values of hyper-parameters $(\alpha, \beta, \gamma, \delta, \epsilon)$ may be required. Say for energy sensitive applications \cite{sarkar2016theoretical, abbas2015survey, kamalinejad2015wireless}, we need to optimize energy even though other metrics might get compromised. Then the loss would have $\alpha=1$ and rest 0. For response time-sensitive applications like healthcare monitoring or traffic management \cite{rahmani2018exploiting}, the loss would have $\beta=1$ and rest 0. Similarly, for different applications, a different set of hyper-parameter values is required. 


Now, for the Neural Network model we need to include the penalty as well because the output described in Section \ref{sec:output} is unconstrained, as done in other works \cite{achiam2017constrained, doshi2016deep}. If we include the penalty defined by Equation \ref{eq:penalty}, then the model updates its parameters to not only minimize $Loss_i$ but also to satisfy constraints described in Equation \ref{eq:action}. Thus, we define the loss for the Neural Network as shown in Equation \ref{eq:loss-fcn}. So,
\begin{equation}
\label{eq:loss-fcn}
\begin{aligned}
Loss_i^{PG} = Loss_i + Penalty_i.
\end{aligned}
\end{equation}

\subsection{Model update}
\begin{figure}[]
    \centering
    \includegraphics[width=\columnwidth]{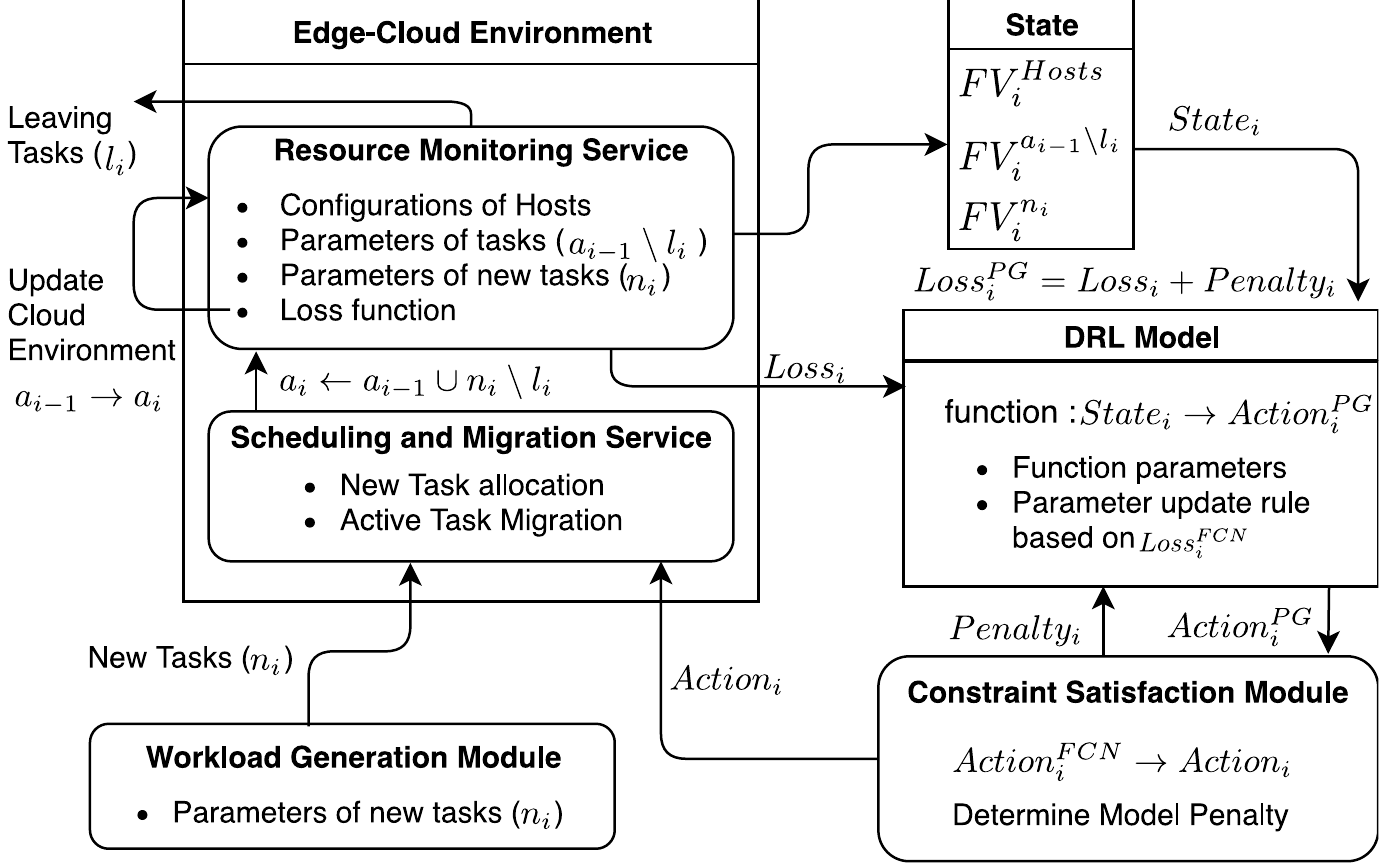}
    \caption{Learning Model}
    \label{fig:model}
\end{figure}

Having defined the input-output specifications and the loss function we now define the procedure to update the Model after every scheduling interval. A summary of the interaction and model update for the transition from interval $SI_{i-1}$ to the interval $SI_i$ is shown in Figure \ref{fig:model}. We consider an episode to contain $n$ scheduling intervals. At the beginning of every scheduling interval say $SI_i$, the WGM sends new tasks to the Scheduling and Migration Service (SMS). Then, SMS and WGM send the $State_i$ to the DRLM which includes the feature vectors of hosts, remaining active tasks from previous interval ($a_{i-1}\setminus l_i$) and new tasks ($n_i$). Also, the RMS sends the $Loss_i$ to the DRLM. The CSM sends $Penalty_i$ based on decision of $Action_{i-1}^{PG}$.The model then generates an $Action_i^{PG}$ and updates its parameters based on Equation \ref{eq:loss-fcn}. which is sent to the CSM. The CSM converts $Action_i^{PG}$ to $Action_i$ and sends it to RMS. It also calculates and stores $Penalty_{i+1}$ for next interval $SI_{i+1}$. The RMS allocates new tasks ($n_i$) and migrates remaining tasks from previous interval ($a_{i-1}\setminus l_i$) based on $Action_i$ received from CSM. This updates $a_{i-1}$ to $a_i$ as $a_i \leftarrow a_{i-1}\cup n_i \setminus l_i$. The tasks in $a_i$ execute for the interval $SI_i$ and the cycle repeats for the next interval $SI_{i+1}$.

\section{Stochastic Dynamic Scheduling using policy gradient learning}
\label{sec:stochastic-dynamic-scheduling}

The complete framework works as follows: at the beginning of every scheduling interval, (1) the RMS receives the task requests including task parameters like computation, bandwidth and SLA requirements. (2) These requirements and the host characteristics from Resource Monitoring Service are used by the DRL model to predict the next scheduling decisions. (3) The constraint satisfaction module finds the possible migration and scheduling decision from the output of DRL model. (4) For the new tasks, the RMS informs the user/IoT device to send its request directly to the corresponding edge/cloud device scheduled for this task. (5)  The loss function is calculated for the DRL model and its parameters are updated. The formulation and the learning model described earlier in Section \ref{sec:learning-model} is generic for any policy based RL model. The model, which is a function form $State_i$ to $Action_i^{PG}$ is assumed to be the theoretically best function for minimizing $Loss_i^{PG}$. There exist many prior works which try to model this function using Q-Table or a neural network function approximator \cite{basu2019Megh, cheng2018drlcloud, zhang2017energydeepQRL} giving a deterministic policy which is unable to adapt in stochastic settings. However, our approach tries to approximate the policy itself and optimize it using policy gradient methods with $Loss_i^{PG}$ as a signal to update the network.

\subsection{Neural Network Architecture}

\begin{figure}[]
    \centering
    \includegraphics[width=\columnwidth]{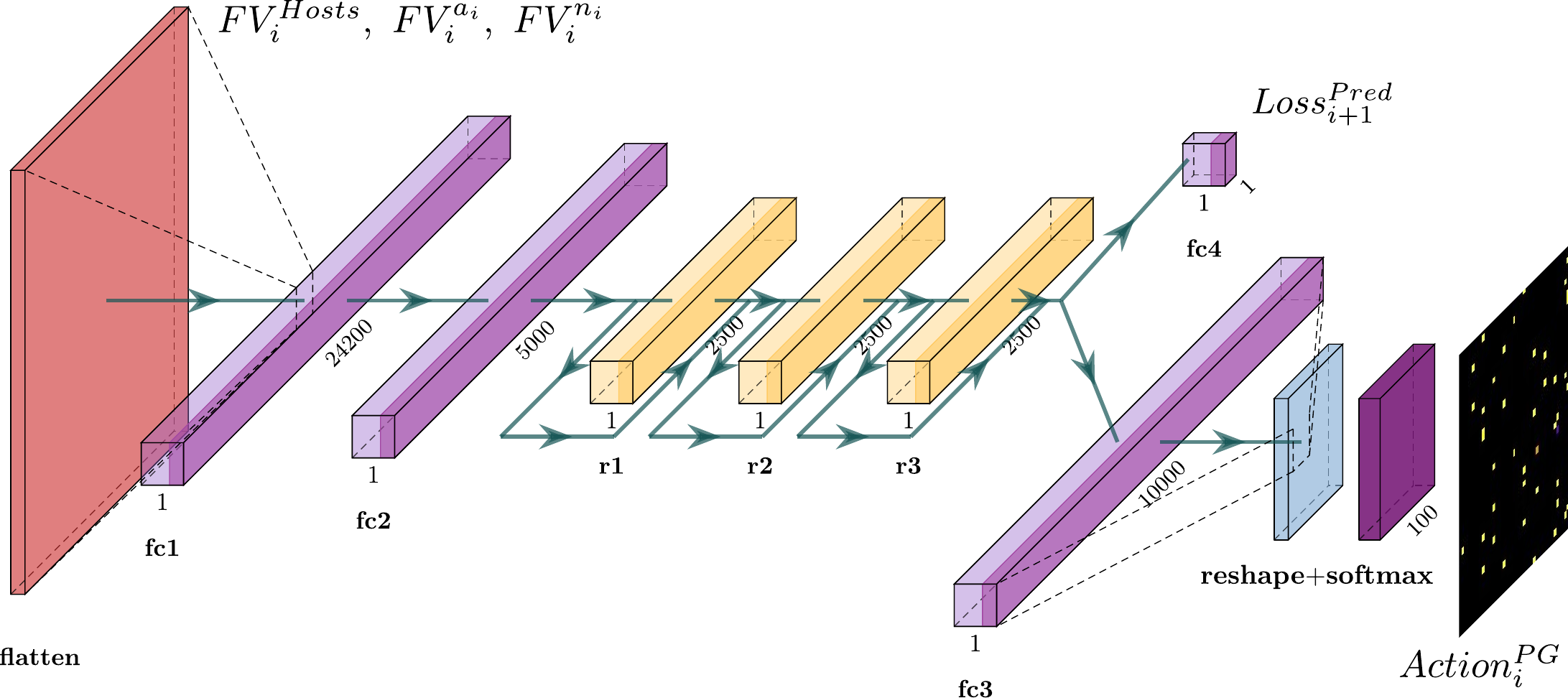}
    \caption{Neural Network Architecture}
    \label{fig:network}
\end{figure}

To approximate the function from $State_i$ to $Action_i^{PG}$ for every interval $SI_i$, we use a R2N2 network. The advantage of using an R2N2 network is its ability to capture complex temporal relationships between the inputs and outputs. The architecture with the layer description used for the proposed work is shown in Figure \ref{fig:network}. A single network is used to predict both policy (actor head) and cumulative loss after the current interval (critic head).

The R2N2 network has 2 fully connected layers followed by 3 recurrent layers with skip connections. A 2-dimensional input is first flattened and then passed through the dense layers. The output of the last recurrent layer is sent to the two network heads. The actor head output is of size $10^4$ which is reshaped to a 2-dimension $100\times100$ vector. This means that the this model can manage maximum 100 tasks and 100 hosts.  This is done for a fair comparison with other methods that have tested on similar settings \cite{beloglazov2012optimal, basu2019Megh}, but for a larger system the network must be changed accordingly. Finally, softmax is applied across the second dimension so that all values are in [0,1] and the sum of all values in a row equals 1. This output (say $O$) can be interpreted as a probability map where $O_{jk}$ represents the probability with which task $T_j^{a_i}$ should be assigned to host $H_k$ which is $k^{th}$ host in an enumeration of $Hosts$. The output of the critic head is a single constant which signifies the value function i.e., the cumulative loss starting from next interval ($CLoss_{i+1}^{PG}$). The recurrent layers are formed using Gated Recurrent Units (GRUs) \cite{dey2017gate}, which model the temporal aspects of the task and host characteristics including tasks' CPU, RAM and bandwidth requirements and hosts' CPU, RAM and bandwidth capacities. Although the GRU layers help in taking an informed scheduling decision by modeling the temporal characteristics, they increase the training complexity due to large number of network parameters. This is solved by using the skip connections between these layers for faster gradient propagation.

\subsection{Pre-processing and Output Conversion}
\label{sec:pre-process-and-output}

The input to the model for the interval $SI_i$ is $State_i$, which is a 2-dimensional vector. This includes $FV_i^{Hosts},\,FV_i^{n_i},\,FV_i^{a_{i-1}\setminus l_i}$. Among these vectors, the values of all elements of the first two are continuous, but the host index in each row of $FV_i^{a_{i-1}\setminus l_i}$ is a categorical value. Hence, the host indices are converted to a one-hot vector of size $n$ and all feature vectors are concatenated. After this, each element in the concatenated vector is normalized based on the minimum and maximum values of each feature and clipped between [0,1]. We denote the feature of element $e$ as $f_e$, and minimum and maximum values for feature $f$ as $min_f$ and $max_f$ respectively. These minimum and maximum values are calculated based on a sample dataset using two heuristic-based scheduling policies: Local-Regression (LR) for task allocation and Maximum-Migration-Time (MMT) for task selection as described in \cite{beloglazov2012optimal}. Then, the feature-wise standardization is done based on Equation \ref{eq:standardization}. Hence, 

\begin{equation}
\label{eq:standardization}
\begin{aligned}
e = 
\begin{cases}
0 \text{ if } max_{f_e} = min_{f_e}\\
min(1, max(0, \frac{e - min_{f_e}}{max_{f_e} - min_{f_e}})) \text{ otherwise}.
\end{cases} \\
\end{aligned}
\end{equation}

This pre-processed input is then sent to the R2N2 model which flattens it and passes through the Dense layers. The output generated $O$ is converted to $Action_i^{PG}$ by first generating the sorted list of host $SortedHosts_i$ with decreasing probability in $O_i$ for all $i$. Then, $Action_i^{PG}(T_k^{m_i\cup n_i}) \leftarrow SortedHosts_k\ \forall\ k \in \{1, 2, ..., |m_i\cup n_i|\}$.

\subsection{Policy Learning}
\label{sec:fcn-learning}

To learn the weights and biases of the R2N2 network, we use the back-propagation algorithm with reward as $-Loss_i^{PG}$. For the current model, we use adaptive learning rate starting from $10^{-2}$ and decrease it to $1/10^{th}$  when the  absolute sum of of change in the reward for the last ten iterations is less than $0.1$. Using reward as $-Loss_i^{PG}$, we perform Automatic Differentiation \cite{paszke2017automatic} to update the network parameters. We accumulate the gradients of local networks at all edge nodes asynchronously and update the global network parameters periodically as described in \cite{mnih2016asynchronous}. The gradient accumulation rule after the $i^{th}$ scheduling interval is given by Equation \ref{eq:grad} similar to the one in \cite{mnih2016asynchronous}. Here $\theta$ denotes the global network parameters and $\theta'$ denotes the local parameters (only one gradient is set because of a single network with two heads). Thus,

\begin{equation}
\label{eq:grad}
\begin{aligned}
d\theta \leftarrow d\theta &- \alpha \nabla_{\theta '} \log [ \pi(State_i; \theta ') ] (Loss_{i}^{PG} + CLoss_{i+1}^{Pred})\\
&+ \alpha \nabla_{\theta '} (Loss_{i}^{PG} + CLoss_{i+1}^{Pred} - CLoss_i^{Pred})^2.
\end{aligned}
\end{equation}

The $log$ term in the Equation \ref{eq:grad} specifies the direction of change in the parameters, $(Loss_{i}^{PG} + CLoss_{i+1}^{Pred})$ term is the predicted cumulative loss in this episode starting from $State_i$. To minimize this, the gradients are proportional to this quantity and have a minus sign to reduce total loss. The second gradient term is the Mean Square Error (MSE) of the predicted cumulative loss with the cumulative loss after one-step look-ahead. The output $Action_i^{PG}$ is converted to $Action_i$ by CSM and sent to the RMS every scheduling interval. Thus, for each interval, there is a forward pass of the R2N2 network. For back-propagation, we use a episode size of 12, thus we save the experience of the previous episode to find and accumulate gradients and update model parameters after 12 intervals. For large batch sizes, parameter updates are slower and for small ones the gradient accumulation is not able to generalize and has high variance. Accordingly, empirical analysis has resulted  into optimal episode size of 12. As described in Section \ref{sec:setup}, the experimental setup has a scheduling interval of  5 minutes, and hence back-propagation is performed every 1 hour of simulation time (after 12 intervals) . 

A summary of the model update and scheduling with back-propagation is shown in Algorithm \ref{alg:scheduling}. To decide the best possible scheduling decision for each scheduling interval, we iteratively pre-process and send the interval state to the R2N2 model with the loss and penalty to update the network parameters. This allows the model to adapt on-the-fly to the environment, user and application specific requirements. 

\begin{algorithm}[t]
\caption{Dynamic Scheduling}
\label{alg:scheduling}
\begin{algorithmic} [1]
\Statex \textbf{Inputs:}
\STATE Number of scheduling intervals $N$
\STATE Batch Size $B$
\Statex \textbf{Begin}
\FOR{interval index $i$ from 1 to $N$}
\IF{$i>1$ and $i\% B == 0$}
\STATE Use $Loss_i^{PG} = Loss_i+Penalty_i$ in RL Model for back-propagation
\ENDIF
\STATE send \textsc{preProcess}($State_i$) to RL Model
\STATE $probabilityMap$ $\leftarrow$ output of RL Model for $State_i$
\STATE ($Action_i$, $Penalty_{i+1}$) $\leftarrow$ \textsc{ConstraintSatisfactionModule}($probabilityMap$)
\STATE Allocate new tasks and migrate existing tasks based on $Action_i$
\STATE Execute tasks in edge-cloud infrastructure for interval $SI_i$
\ENDFOR{}
\Statex \textbf{End}
\end{algorithmic}
\end{algorithm}

\textit{Complexity Analysis:} The complexity of Algorithm \ref{alg:scheduling} depends on multiple tasks. The pre-processing of the input state is $O(ab)$ where $a\times b$ is the maximum size of feature vector among the vectors $FV_i^{Hosts},\,FV_i^{n_i},\,FV_i^{a_{i-1}\setminus l_i}$. To generate the $Action_i$ and $Penalty_i$ the CSM takes  $O(n^2)$ time for $n$ hosts and tasks based on Equations \ref{eq:penalty} and \ref{eq:action-selection}. As the feature vectors have a higher cardinality than the number of hosts or tasks, $O(ab)$ dominates $O(n^2)$. Therefore, discarding the forward pass and back-propagation (as they are performed in Graphics Processing Units - GPU \cite{li2014large}), for $N$ scheduling intervals, the total time complexity is $O(abN)$.

\section{Performance Evaluation}

\begin{table*}[!t]
    \centering
    \resizebox{\textwidth}{!}{
\begin{tabular}{|c|c|c|c|c|c|c|c|c|c|c|c|c|c|c|c|c|c|c|}
\hline 
\multirow{2}{*}{Name} & \multirow{2}{*}{Processor} & Core & \multirow{2}{*}{MIPS} & \multirow{2}{*}{RAM} & Network & Disk & Cost & \multicolumn{11}{c|}{SPEC Power (Watts) for different CPU percentage usages}\tabularnewline
\cline{9-19} 
 &  & count &  &  & Bandwidth & Bandwidth & Model & 0\% & 10\% & 20\% & 30\% & 40\% & 50\% & 60\% & 70\% & 80\% & 90\% & 100\%\tabularnewline
\hline 
\hline 
\multicolumn{19}{|c|}{Edge Layer}\tabularnewline
\hline 
\hline 
Hitachi HA 8000 & Intel i3 3.0 GHz & 2 & 1800 & 8 GB & 0.1 GB/s & 76 MB/s & 0.11 \$/hr & 24.3 & 30.4 & 33.7 & 36.6 & 39.6 & 42.2 & 45.6 & 51.8 & 55.7 & 60.8 & 63.2\tabularnewline
\hline 
DEPO Race X340H & Intel i5 3.2 GHz & 4 & 2000 & 16 GB & 1 GB/s & 49 MB/s & 0.23 \$/hr & 83.2 & 88.2 & 94.3 & 101 & 107 & 112 & 117 & 120 & 124 & 128 & 131\tabularnewline
\hline 
\hline 
\multicolumn{19}{|c|}{Cloud Layer}\tabularnewline
\hline 
\hline 
Dell PowerEdge R820 & Intel Xeon 2.6 GHz & 32 & 2000 & 48 GB & 1 GB/s & 49 MB/s & 3.47 \$/hr & 110  & 149 & 167 & 188 & 218 & 237 & 268 & 307 & 358 & 414 & 446\tabularnewline
\hline 
Dell PowerEdge C6320 & Intel Xeon 2.3 GHz & 64 & 2660 & 64 GB & 1.5 GB/s & 1024 MB/s & 6.94 \$/hr & 210 & 371 & 449 & 522 & 589 & 647 & 705 & 802 & 924 & 1071 & 1229\tabularnewline
\hline 
\end{tabular}}
    \caption{Configuration of Hosts in the Experiment Set Up}
    \label{tab:hosts}
\end{table*}

In this section, we describe the experimental set up, evaluation metrics, dataset and give a detailed analysis of results comparing our model with several baseline algorithms.
\label{sec:perf-eval}

\subsection{Experimental Set Up}
\label{sec:setup}

To evaluate the proposed Deep Learning-based scheduling framework, we developed a simulation environment by extending the elements of iFogSim \cite{gupta2017ifogsim} and  CloudSim toolkits \cite{calheiros2011cloudsim} which already have resource monitoring services inbuilt. As described in Section \ref{sec:fcn-learning}, the execution of the simulation was divided into equal-length scheduling intervals. The interval size was chosen to be 5 minutes long, same as in other works \cite{beloglazov2012optimal, basu2019Megh, cheng2018drlcloud} for a fair comparison with baseline algorithms. The tasks, named as Cloudlets in iFogSim nomenclature, are generated by the WGM based on Bitbrain dataset \cite{shen2015statisticalBitBrain}. We extended the modules of iFogSim and CloudSim to allow the use of parameters like response time, cost and power of edge nodes. We also created new modules to simulate mobility of IoT devices using bandwidth variations, delayed execution of tasks and interact with deep learning software. Additional software for Constraint Satisfaction Module, input pre-processing and output conversion was developed.

The loss function is calculated based on host and task monitoring services in CloudSim. The penalty is calculated by the CSM and sent to the DRLM for model parameter update. We now describe in more detail the dataset, task generation and duration implementation, hosts' configuration and metrics for evaluation.


\subsubsection{Dataset}

In the simulation environment, the tasks (cloudlets) are assigned to Virtual Machines (VMs) which are then allocated to hosts. For the current setting of task on edge-cloud environment, we consider a bijection from cloudlets to VMs by allocating $i^{th}$ created Cloudlet to $i^{th}$ created VM and discard the VM when the corresponding Cloudlet is completed. The dynamic workload is generated for cloudlets based on real-world open-source Bitbrain's dataset \cite{shen2015statisticalBitBrain}\footnote{The BitBrain dataset can be downloaded from: 
\url{http://gwa.ewi.tudelft.nl/datasets/gwa-t-12-bitbrains}}.

\begin{figure}[!b]
    \centering
    \subfigure[CPU and RAM characteristics]{
    \includegraphics[width=.48\columnwidth]{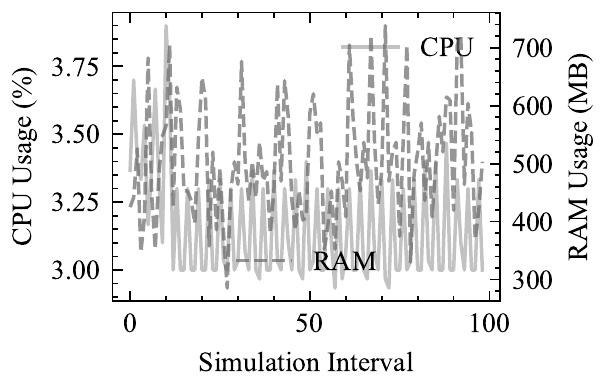}
    \label{fig:cpu-ram}
    }
    \subfigure[Disk and Network Bandwidth characteristics]{
    \includegraphics[width=.45\columnwidth]{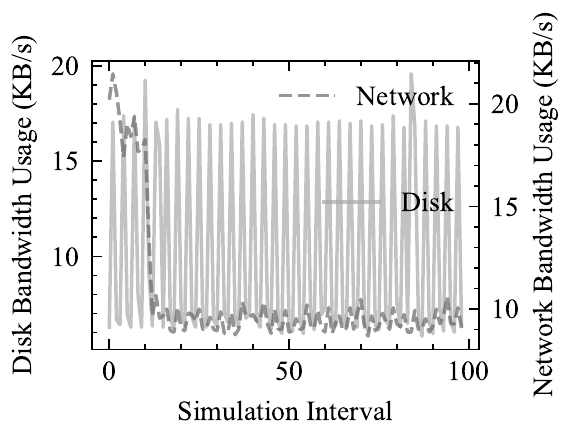}
    \label{fig:disk-nw}
    }
    \caption{Bitbrain Dataset Characteristics}
    \label{fig:dataset}
\end{figure}

The Bitbrain's dataset \cite{shen2015statisticalBitBrain} has real traces of resource consumption metrics of business-critical workload hosted on Bitbrain infrastructure. This data includes logs of over 1000 VMs workload hosting on two types of machines. We have chosen this dataset as it represents real-world infrastructure usage patterns, which is useful to construct precise input feature vectors for learning models. The dataset consists of workload information for each time-stamp (separated by 5 minutes) including the number of requested CPU cores, CPU usage in terms of MIPS, RAM requested with Network (receive/transmit) and Disk (read/write) bandwidth characteristics. These different categories of workload data constitute the feature values of $FV_i^{n_i}$ and $FV_i^{a_{i-1}\setminus l_i}$, where the latter also has an index of host allocated in the previous scheduling/simulation interval. The CPU, RAM, network bandwidth and disk characteristics for a random node and its trace in the BitBrain dataset are shown to be highly volatile in Figure \ref{fig:dataset}.

We divide the dataset into two partitions of $25\%$ and $75\%$ VM workloads. The larger partition is used for training of the R2N2 network and the former partition is used for testing of the network, sensitivity analysis and comparison with other related works.

\subsubsection{Task generation and duration configuration}

In the proposed work, we consider a dynamic task generation model. Prior work \cite{beloglazov2012optimal} does not consider a dynamic task generation environment, which is not close to the real-world setting. At the beginning of every interval, the WGM sends $n_i$ new tasks where $|n_i|$ is normal distributed $\mathcal{N}(\mu_{n},\sigma_{n}^2)$. Also, each task $t \in n_i$ has an execution duration of $\mathcal{N}(\mu_{t},\sigma_{t}^2)$ seconds. In our setting, we kept 100 hosts and no more than 100 tasks in the system being scheduled on 10 actor-agents (schedulers). We keep in our simulation environment: $(\mu_{n_i},\sigma_{n_i}) = (12, 5)$ and $(\mu_{t},\sigma_{t}) = (1800, 300)\, seconds$ for number of new tasks and duration of tasks respectively. At the time of task creation, for already active $|a_{i-1}\setminus l_i|$ tasks, we only create $min(100-|a_{i-1}\setminus l_i|, \mathcal{N}(\mu_{n_i},\sigma_{n_i}^2))$ tasks so that $|a_i|$ does not exceed 100. This limit is required because the size of the input to the R2N2 network has a prefixed upper limit which in our case is 100.

\subsubsection{Hosts - Edge and Cloud nodes}

The infrastructure considered in our studies is a heterogeneous edge-cloud based environment. Unlike prior work \cite{cheng2018drlcloud, basu2019Megh, mao2016RMSDRL,  zhang2017energydeepQRL}, we consider both resource-constrained edge-cloud devices closer to the user and thus having lower response time and also resource-abundant cloud nodes with much higher response time. In our settings, we have considered response time of edge-cloud nodes to be 1 ms and that of cloud nodes to be 10 ms based on the empirical studies using the \textit{Ping} utility in an existing edge-cloud framework namely FogBus \cite{TULI201922}. 

Moreover, the environment considered is heterogeneous with a diverse range of computation capabilities of edge and cloud host. A summary of CPU, RAM, Network and other capacities with the Cost Model is given in Table \ref{tab:hosts}, 25 instances of each host type in the environment. The cost model for the cloud layer is based on Microsoft Azure IaaS cloud service. The cost per hour (in US Dollar) is calculated based on the costs of similar configuration machines offered by Microsoft Azure in South-East Australia\footnote{Microsoft Azure pricing calculator for South-East Australia \url{https://azure.microsoft.com/en-au/pricing/calculator/}}. For the edge nodes, the cost is based on the energy consumed by the edge node. As per the targeted environment convention, we choose resource-constrained machines at edge (Intel i3 and Intel i5) and powerful rack server as cloud nodes (Intel Xeon).  The power consumption averaged over the different SPEC benchmarks \cite{spec} for respective machines is shown in Table \ref{tab:hosts}. However, the power consumption values shown in Table \ref{tab:hosts} are average values over this specific benchmark suite. Power consumption of hosts also depends on RAM, Disk and bandwidth consumption characteristics and are provided to the model by the underlying CloudSim simulator. In the execution environment, we consider the host capacities (CPU, RAM, Network Bandwidth, etc) and the current usage to form the feature vector $FV_i^{Hosts}$ for the $i^{th}$ scheduling interval. For the experiments, we keep the testing simulation duration of  1 day, which equals to total 288 scheduling intervals.

\subsection{Evaluation Metrics}
\label{sec:metrics}
To evaluate the efficacy of the proposed A3C-R2N2 based scheduler, we consider the following metrics. \black{Motivated from prior works~\cite{basu2019Megh, tuli2020healthfog, TULI201922}, energy is paramount in resource constrained edge-cloud environments and real-time tasks require low response times. Moreover, service level agreements are crucial in time-critical tasks and low execution cost is required for budget task execution.}:
\begin{enumerate}
    \item \textit{Total Energy Consumption} which is given as $\sum_{h\in Hosts} \int_{t = t_{i}}^{t_{i+1}} P_h(t)dt$ for the complete simulation duration. 
    \item \textit{Average Response Time} which is given as $\frac{\sum_{t \in l_{i+1}} Response\ Time(t)}{|l_{i+1}|}$.
    \item \textit{SLA Violations} which is given as  $\frac{\sum_i SLAV_i \cdot |l_{i+1}|}{\sum_i l_i}$ where $SLAV_i$ is defined by Equation \ref{eq:sla}.
    \item \textit{Total Cost} which is given as $\sum_i \sum_{h\in Hosts} \int_{t = t_{i}}^{t_{i+1}} C_h(t)dt$.
\end{enumerate}
Other metrics of importance include: \textit{Average Task Completion Time}, \textit{Total number of completed Tasks} with fraction of tasks that were completed within the expected execution time (based on requested MIPS), \textit{Number of task migrations} in each interval and \textit{Total migration time} per interval. The task completion time is defined as the sum of the average task scheduling time, task execution time and response time of host on which the task ran in last scheduling interval.

\begin{figure*}
    \centering
    \subfigure[Total Energy Consumption]{
    \includegraphics[width=.23\textwidth]{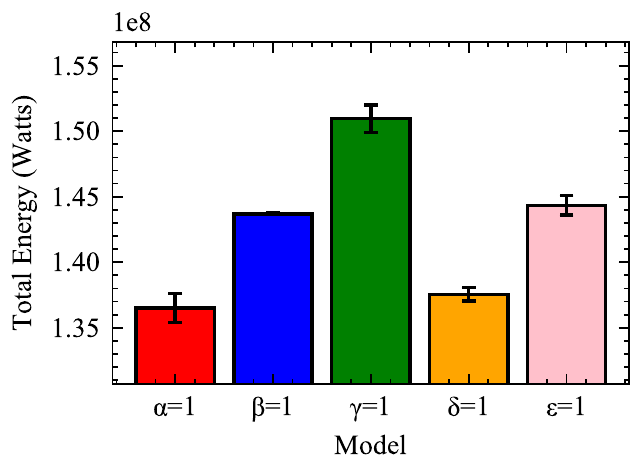}
    \label{fig:loss-energy}
    }
    \subfigure[Average Response Time]{
    \includegraphics[width=.23\textwidth]{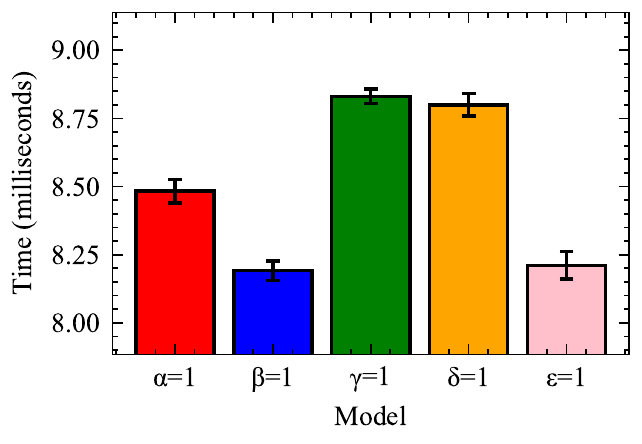}
    \label{fig:loss-response}
    }
    \subfigure[Fraction of SLA Violations]{
    \includegraphics[width=.23\textwidth]{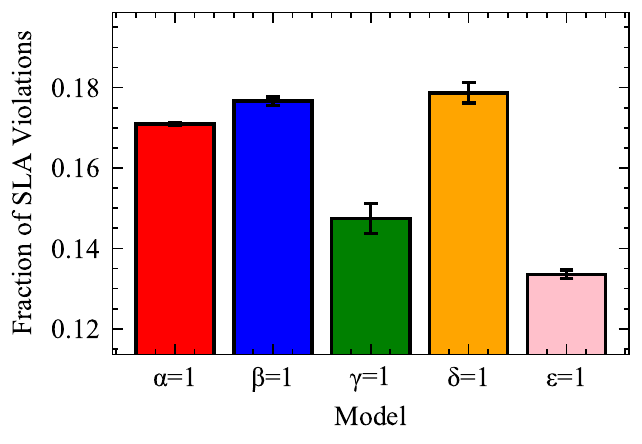}
    \label{fig:loss-sla}
    }
    \subfigure[Total cost]{
    \includegraphics[width=.23\textwidth]{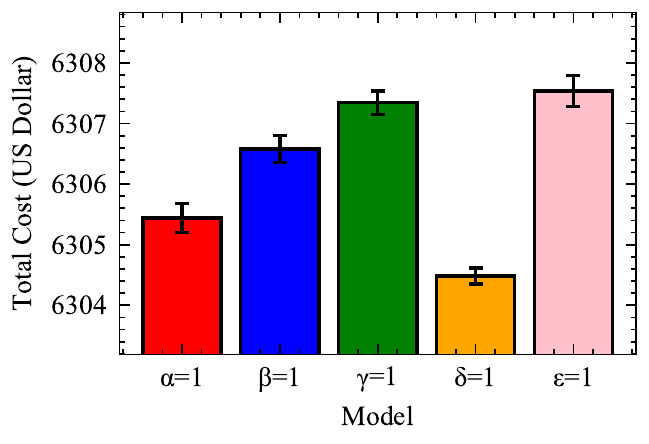}
    \label{fig:loss-cost}
    }\\
    \subfigure[Average Task Completion Time]{
    \includegraphics[width=.22\textwidth]{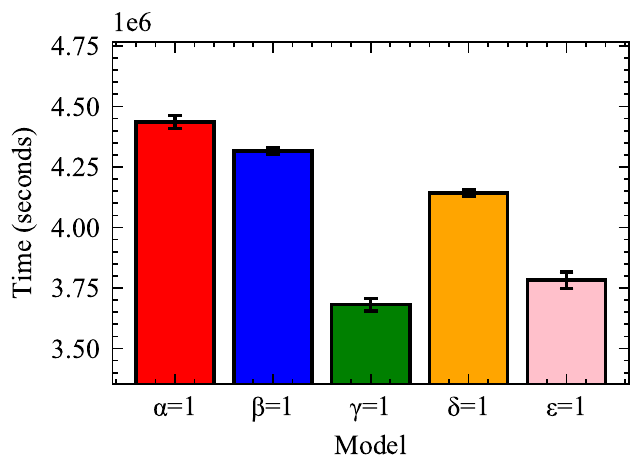}
    \label{fig:loss-completion}
    }
    \subfigure[Number of total completed tasks]{
    \includegraphics[width=.24\textwidth]{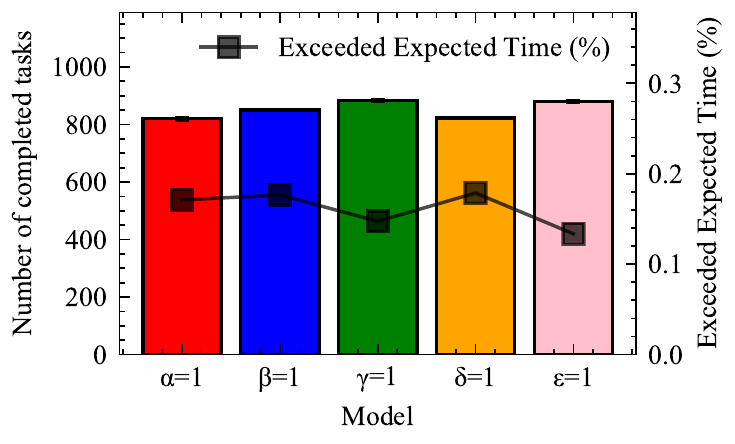}
    \label{fig:loss-tasks}
    }
    \subfigure[Number of task migration in each interval]{
    \includegraphics[width=.22\textwidth]{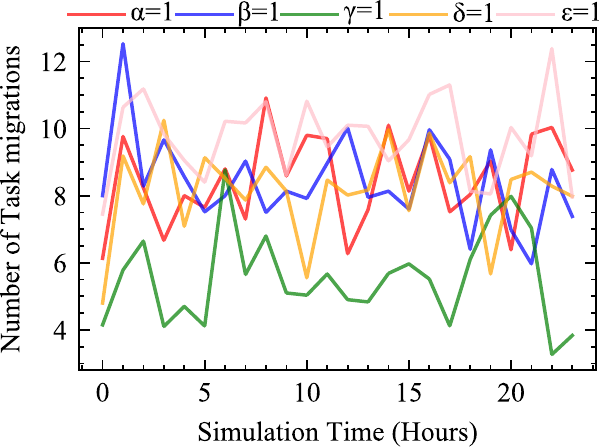}
    \label{fig:loss-migrations}
    }
    \subfigure[Total Migration Time in each interval]{
    \includegraphics[width=.22\textwidth]{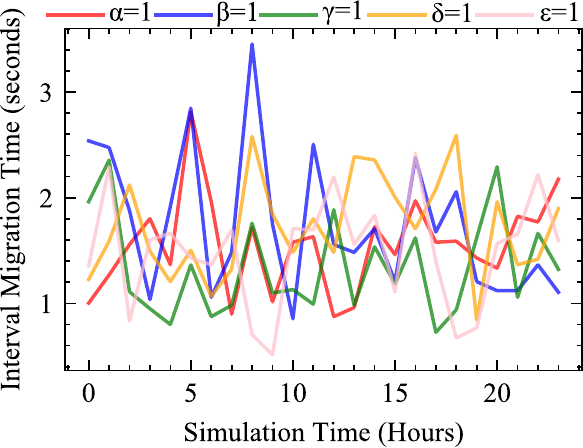}
    \label{fig:loss-migration-time}
    }
    \caption{Comparison of Model Trained with Different Loss Functions}
    \label{fig:loss-comparison}
\end{figure*}

\subsection{Baseline Algorithms}
\label{sec:baseline-algos}
We evaluate the performance of our proposed algorithms with the following baseline algorithms, \black{the reasons for choosing these is described in Section~\ref{sec:related-work}}. Multiple heuristics have been proposed by  \cite{beloglazov2012optimal} for dynamic scheduling. These are a combination of different sub heuristics for different sub-problems such as host overload detection and task/VM selection and we have selected the best three heuristics from those. All of these variants use \textit{Best Fit Decreasing (BFD)} heuristics to identify the target host. Furthermore,  we also compare our results to two types of standard \textit{RL} approaches that are widely used in the literature.

\begin{itemize}[leftmargin=*]
\item \textit{LR-MMT:}  schedules workloads dynamically based on \textit{Local Regression (LR)}  and \textit{Minimum Migration Time (MMT)} heuristics for overload detection and task selection, respectively (details in \cite{beloglazov2012optimal})
\item \textit{MAD-MC:} schedules workloads dynamically based on \textit{Median Absolute Deviation (MAD)}  and \textit{Maximum Correlation Policy (MC)} heuristics for overload detection and task selection, respectively (details in \cite{beloglazov2012optimal})
\item \textit{DDQN:} standard \textit{Deep Q-Learning} based \textit{RL} approach, many works have used this technique in literature including \cite{basu2019Megh, mao2016RMSDRL, zhang2017energydeepQRL}.  We implement the optimized Double DQN technique.
\item \textit{DRL (REINFORCE):} policy gradient based REINFORCE method with fully connected neural network \cite{mao2016resource}.
\end{itemize}
 It is important to note that we implement these algorithms adapting to our problem and compare the results. The RL model that has been used for comparison with our proposed model uses a state representation same as the $State_i$ defined in Section \ref{sec:input} for fair comparison. An action is a change from one state to another in the state space. As in \cite{cheng2018drlcloud}, the DQN network is updated using Bellman Equation \cite{watkins1992q} with the reward defined as $-Loss_i^{PG}$. The REINFORCE method is implemented without asynchronous updates or recurrent network.

\subsection{Analysis of Results}

In this subsection, we provide the experimental results using the experimental setup and the dataset described in Section \ref{sec:setup}. We also discuss and compare our results based on evaluation metrics specified in Section \ref{sec:metrics}. We first analyze the sensitivity of hyper-parameters $(\alpha, \beta, \gamma, \delta, \epsilon)$ on the model learning and how it affects different metrics. We then analyze the variation of scheduling decisions based on different hyper-parameter values and show how the combined optimization of different evaluation metrics provides better results. We also compare the fraction of scheduling time with total execution time by varying the number of layers on the R2N2 network. Based on the above analysis, we find the optimum R2N2 network and hyper-parameter values to compare with the baseline algorithms described in Section \ref{sec:baseline-algos}. All model learning is done for 10 days of simulation time and testing is done for 1 day of simulation time using a disjoint set of workloads of the dataset.

\subsubsection{Sensitivity Analysis of Hyper-parameters}

We first provide experimental results in Figure \ref{fig:loss-comparison} for different hyper-parameter values and show how changing the loss function to learn only one of the metric of interest specifically, varies the learned network to give different values of the evaluation metrics, these experiments were carried for a single day of simulation duration. To visualize the output probability map from the R2N2 network, we display it using a color map to depict probabilities (0 to 1) of allocating tasks to hosts as described in Section \ref{sec:pre-process-and-output}.

\begin{figure}[]
    \centering
    \includegraphics[width=.6\columnwidth]{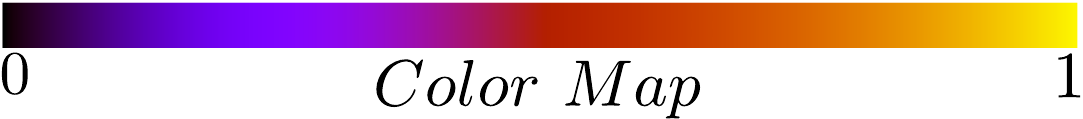}
    \\
    \subfigure[EMN]{
    \includegraphics[width=.3\columnwidth]{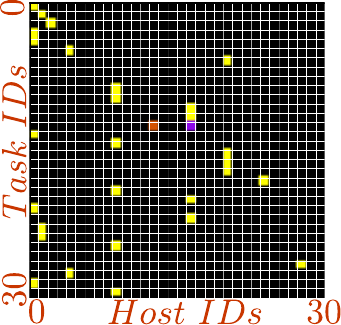}
    \label{fig:energy-map}
    }
    \subfigure[CMN]{
    \includegraphics[width=.3\columnwidth]{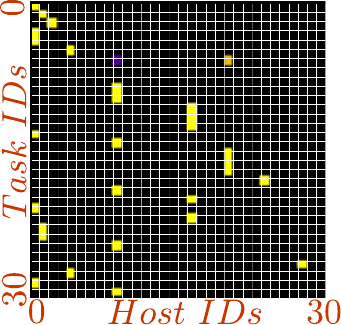}
    \label{fig:cost-map}
    }
    \caption{Probability Map for EMN and CMN showing similarity and positive correlation}
    \label{fig:map-energy-cost}
\end{figure}

When $\alpha = 1$ (rest = 0), then the R2N2 network solely tries to optimize the average energy consumption, and hence we call it \textit{Energy Minimizing Network} (EMN). The total energy consumed across the simulation duration is least for this network as shown in Figure \ref{fig:loss-energy}. As low energy devices (edge nodes) consume the least energy and also have least cost, energy is highly correlated to cost, and hence the \textit{Cost Minimizing Network} (CMN, $\delta = 1$) also has very low total energy consumption. As shown in Figure \ref{fig:map-energy-cost}, for the same $State_i$, the probability map and hence the allocation are similar for both networks. Similarly, we can also see that in Figure \ref{fig:loss-cost}, CMN has the least cost and the next least cost is achieved by EMN.

\begin{figure}[!t]
    \centering
    \subfigure[RTMN]{
    \includegraphics[height=.45\columnwidth]{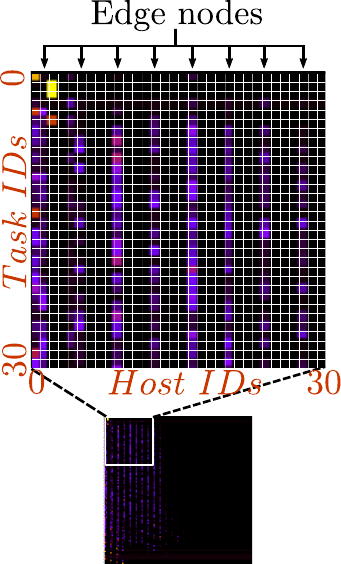}
    \label{fig:response-map}
    }
    \subfigure[SLAVMN]{
    \includegraphics[height=.45\columnwidth]{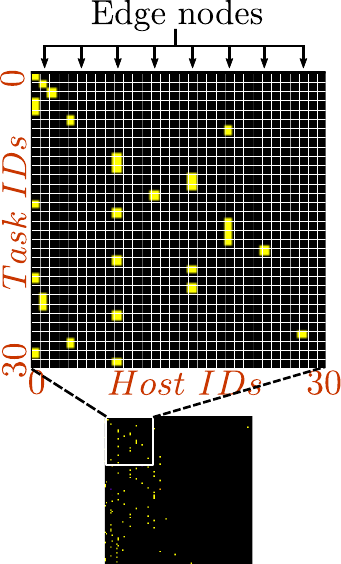}
    \label{fig:sla-map}
    }
    \caption{Probability Map for RTMN and SLAVMN showing that the former does not distinguish among edge nodes but SLAVMN does}
    \label{fig:map-response-sla}
\end{figure}

The graph in Figure \ref{fig:loss-response} shows that the \textit{Response Time Minimizing Network} (RTMN, $\beta = 1$) has the least average response time and tries to place most of the tasks on edge nodes also shown in Figure \ref{fig:response-map}. Moreover, this network does not differentiate among the edge nodes in terms of their CPU loads because all edge nodes have the same response time and hence gives almost same probability to every edge node for each task. The \textit{SLA Violation Minimizing Network} (SLAVMN, $\epsilon = 1$) also has a low response time as a number of SLA violations are directly related to response time for tasks. However, SLA violations also depend on the completion time of tasks, and as the average task completion time of RTMN is very high, the SLA violations of this network are much more than the other network as shown in Figure \ref{fig:loss-sla}. The fraction of SLA violation is least for SLAVMN and next least is for the \textit{Migration Time Minimizing Network} (MMN, $\gamma = 1$). The SLAVMN network also sends tasks to edge nodes like RTMN, but it also considers task execution time and CPU loads to distribute tasks more evenly as shown in Figure \ref{fig:sla-map}.


\begin{figure}
\centering
\begin{minipage}{.48\columnwidth}
  \centering
    \includegraphics[width=\linewidth]{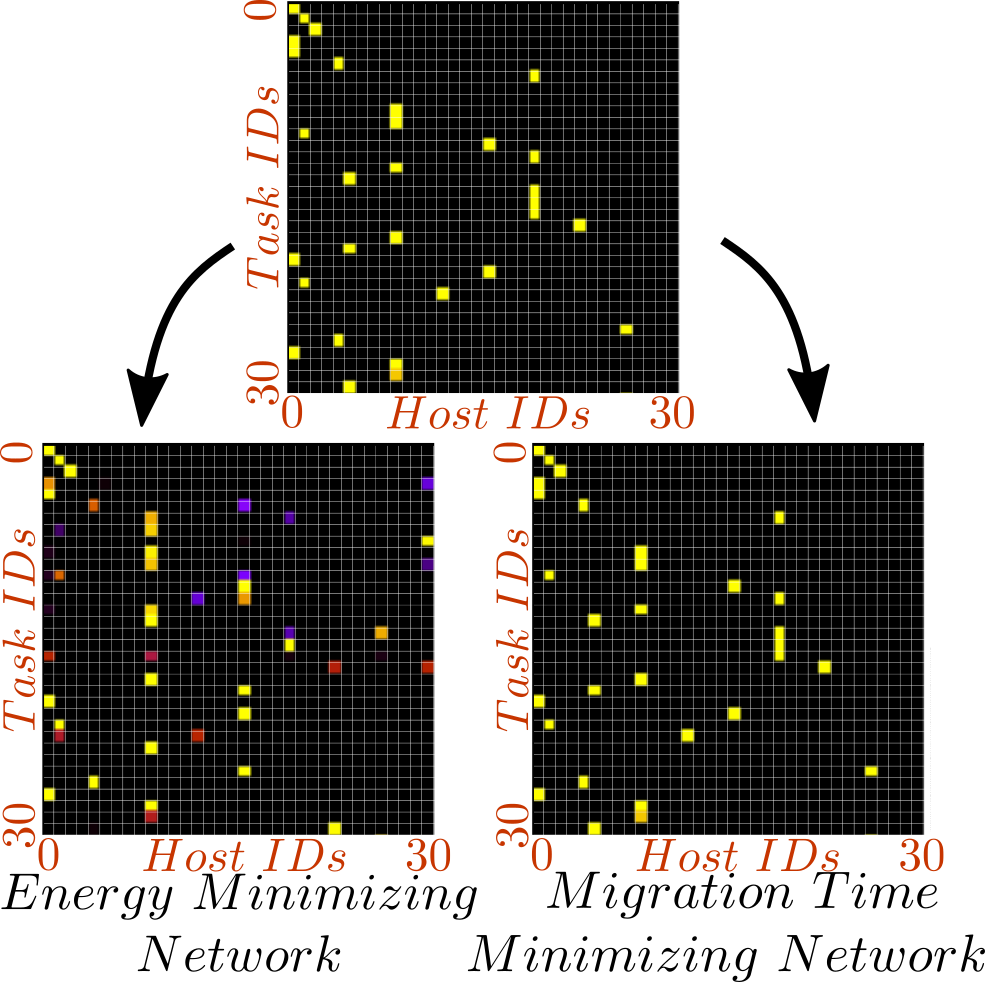}
    \caption{Probability Maps showing that MTMN has lesser migrations than EMN}
    \label{fig:map-energy-mig}
\end{minipage}
\hfill
\begin{minipage}{.49\columnwidth}
  \centering
    \includegraphics[width=\linewidth]{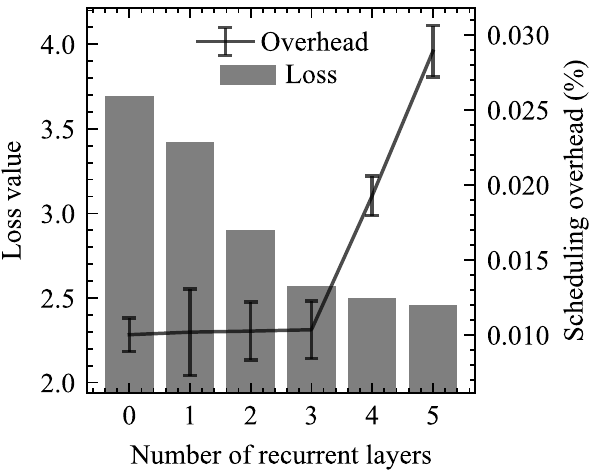}
    \caption{Loss and scheduling overhead with number of \black{recurrent} layers}
    \label{fig:layers}
\end{minipage}
\end{figure}

When only average migration time is being optimized, the average task completion time is minimum, as shown in Figure \ref{fig:loss-completion}. However,  the SLA violation is not minimum as this network does not try to minimize the response time of tasks, as shown in Figure \ref{fig:loss-response}. Moreover, the number of completed tasks is highest for this network as shown in Figure \ref{fig:loss-tasks}. Still, the fraction of tasks completed within the expected time is highest for SLAVMN. Figures \ref{fig:loss-migrations} and \ref{fig:loss-migration-time} show that number of task migrations and migration time is least for MTMN. Also compared in Figure \ref{fig:map-energy-mig} the number of migrations for the sample size of 30 initial tasks are 7 for EMN and 0 for the other.



\begin{figure}{!t}
\centering
\begin{minipage}{.48\columnwidth}
  \centering
  \includegraphics[width=\linewidth]{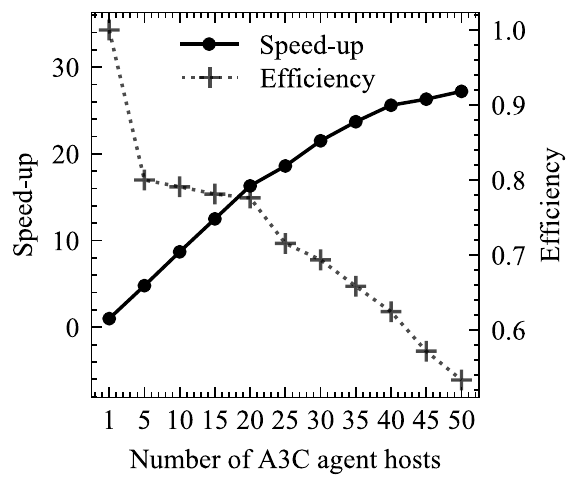}
    \caption{Scalability of A3C-R2N2}
    \label{fig:scale}
\end{minipage}%
\begin{minipage}{.5\columnwidth}
  \centering
  \includegraphics[width=\linewidth]{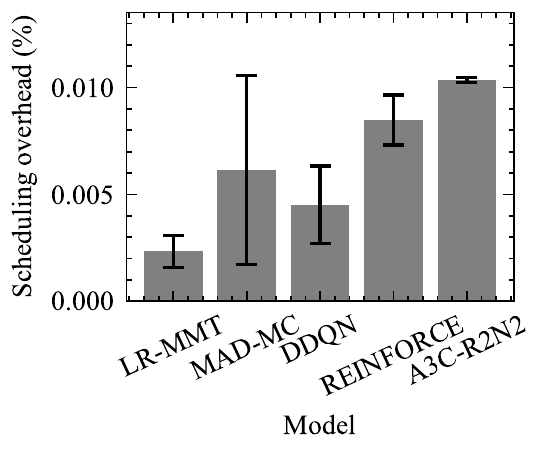}
    \caption{Overheads}
    \label{fig:overhead}
\end{minipage}
\end{figure}

\begin{figure*}[!t]
    \centering
    \subfigure[Total Energy Consumption]{
    \includegraphics[width=.23\textwidth]{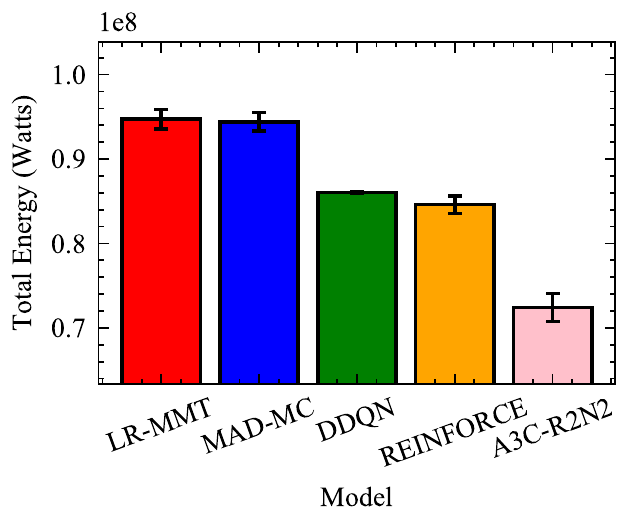}
    \label{fig:fcn-energy}
    }
    \subfigure[Average Response Time]{
    \includegraphics[width=.23\textwidth]{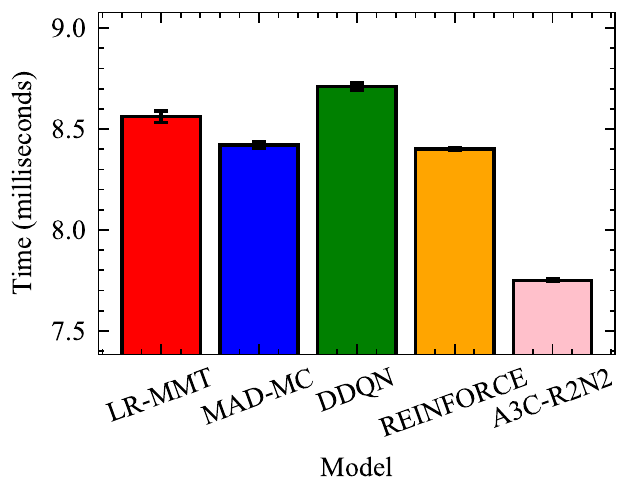}
    \label{fig:fcn-response}
    }
    \subfigure[Fraction of SLA Violations]{
    \includegraphics[width=.23\textwidth]{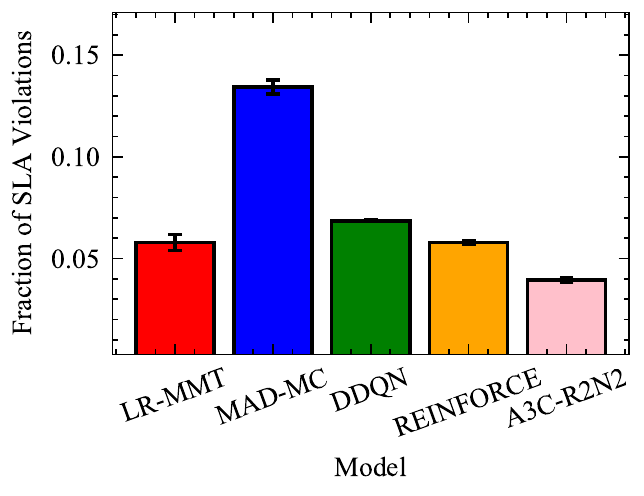}
    \label{fig:fcn-sla}
    }
    \subfigure[Total cost]{
    \includegraphics[width=.23\textwidth]{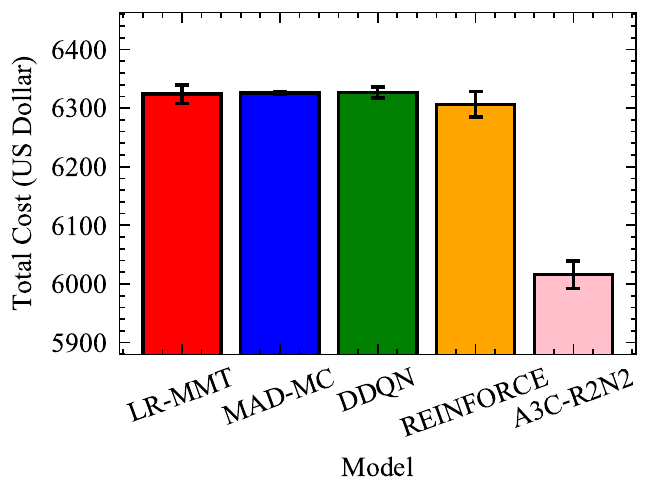}
    \label{fig:fcn-cost}
    }\\
    \subfigure[Average Task Completion Time]{
    \includegraphics[width=.22\textwidth]{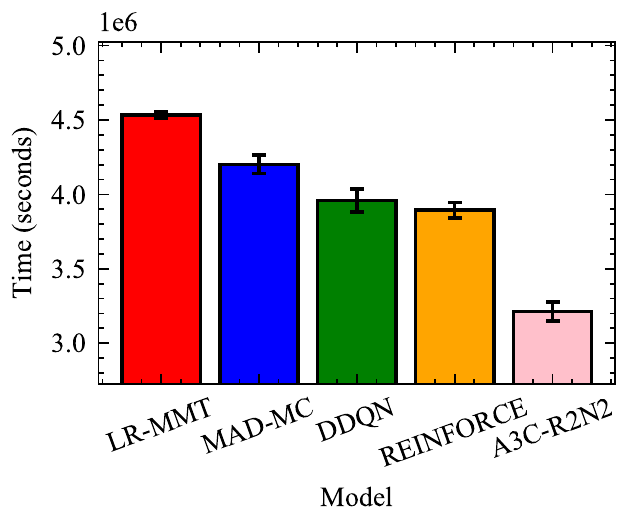}
    \label{fig:fcn-completion}
    }
    \subfigure[Number of total completed tasks]{
    \includegraphics[width=.24\textwidth]{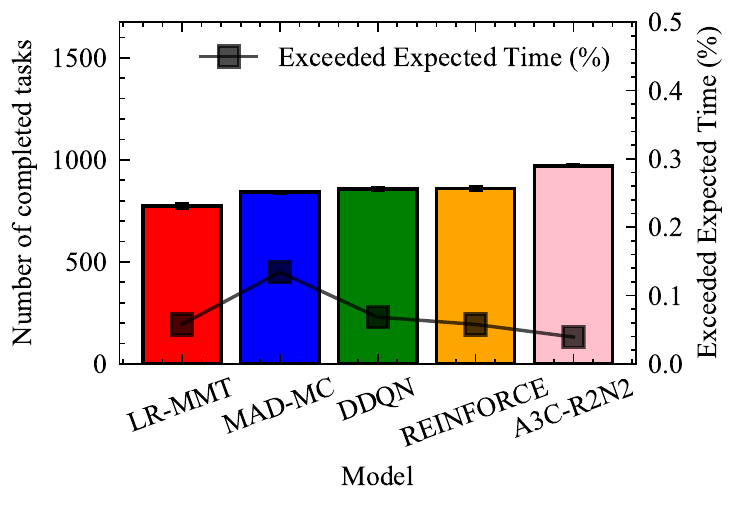}
    \label{fig:fcn-tasks}
    }
    \subfigure[Number of task migration in each interval]{
    \includegraphics[width=.22\textwidth]{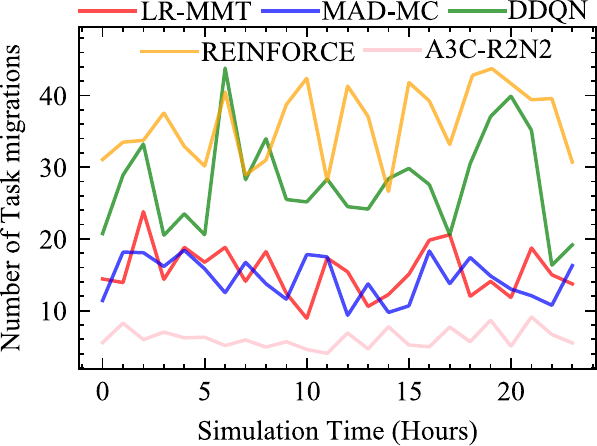}
    \label{fig:fcn-migrations}
    }
    \subfigure[Total Migration Time in each interval]{
    \includegraphics[width=.22\textwidth]{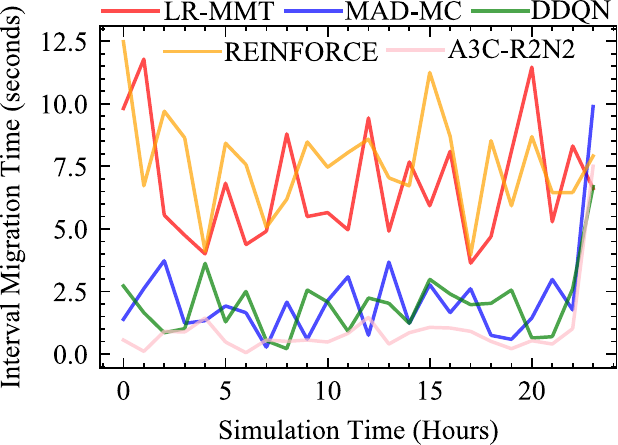}
    \label{fig:fcn-migration-time}
    }
    \caption{Comparison of Deep Learning Model with prior Heuristic-based Works}
    \label{fig:fcn}
\end{figure*}

Optimizing each of the evaluation metrics independently shows that the R2N2 based network can adapt and update its parameters to learn the dependence among tasks and hosts to reduce metric of interest which may be energy, response time, etc. However, for the optimum network, we use a combination of all metrics. This combined optimization leads to a much lower value of the loss and a much better network. This is because optimizing only along one variable might reach a local optimum and the loss of hyper-parameter space being a highly non-linear function, combined optimization leads to much better network \cite{miettinen2012nonlinear}. Based on the empirical evaluation for each combination and block coordinate descent \cite{wright2015coordinate} for minimizing $Loss$, the optimum values of the hyper-parameters are given by Equation \ref{eq:optimum-hyper-params}. Thus,

\begin{equation}
    \begin{aligned}
    (\alpha,\beta,\gamma,\delta,\epsilon) = (0.4, 0.16, 0.174, 0.135, 0.19).
    \end{aligned}
    \label{eq:optimum-hyper-params}
\end{equation}



\subsubsection{Sensitivity Analysis of the number of layers}

Now that we have the optimum values of hyper-parameters, we analyze the scheduling overhead with the number of recurrent layers of the R2N2 network. The scheduling overhead is calculated as the ratio of time taken for scheduling to the total execution duration in terms of simulation time. As shown in Figure \ref{fig:layers}, the value of the loss function decreases with the increase in the number of layers of the Neural Network. This is expected because as the number of layers increase so do the number of parameters and thus the ability of the network to fit more complex functions becomes better. The scheduling overhead depends on the system on which the simulation is run, and for the current experiments, the system used had CPU - Intel i7-7700K and GPU - Nvidia GTX 1070 graphics card (8GB graphics RAM). As shown in the figure, there is an inflection point at 3 recurrent layers because the R2N2 network with 4 or more such layers could not fit in the GPU graphics RAM. Based on the available simulation infrastructure, for the comparison with baseline algorithms, we use the R2N2 network with 3 recurrent layers and hyper-parameter values given by Equation \ref{eq:optimum-hyper-params}.

\subsubsection{Scalability Analysis}

We now show how the A3C-R2N2 model scales with the number of actor agent hosts in the setup. As discussed in Section~\ref{sec:system_model}, we have multiple edge-cloud nodes in the environment which run the policy learning as described in Section~\ref{sec:fcn-learning}. However, the number of such agents affects the time to train the Actor-Critic network. We define the time taken by $n$ agents to reduce the loss value to $2.5$ as $Time_n$. Now, speedup corresponding to a system with $n$ actors is calculated as $S_n = \frac{Time_1}{Time_n}$. Moreover, efficiency of a system with $n$ agents is defined as $E_n = \frac{S_n}{n}$\cite{eager1989speedup}. Figure~\ref{fig:scale} shows how speedup and efficiency of the model vary with number of agent nodes. As shown, the speedup increases with $n$, however, efficiency reduces as $n$ increases in a piece-wise linear fashion. There is a sudden drop in efficiency when number of agents is increased from 1. This is because of the communication delay between agents which leads to slower model updates. The drop increases again after 20 hosts due to addition of GPU-less agents after 20 hosts. Thus, having agent run only on CPU significantly reduces the efficiency of the proposed architecture. For our experiments, we keep all active edge-cloud hosts (100 in our case) as actor agents in the A3C learning for faster convergence and worst-case overhead comparison. In such a case, the speedup is 34.3 and efficiency is 0.37.

\subsubsection{Evaluation with Baseline Algorithms}

Having the empirically best set of values of hyper-parameters and the number of layers and discussed the scalability aspects of the model, we now compare our policy gradient based reinforcement learning model with the baseline algorithms described in Section \ref{sec:baseline-algos}. The graphs in Figure \ref{fig:fcn} provide results for 1 day of simulation time with a scheduling interval of 5 minutes on the Bitbrain dataset.

Figure \ref{fig:fcn-energy} shows that among the baseline algorithms, DDQN and REINFORCE have the least energy consumption, but A3C-R2N2 model has even lower energy consumption which is 14.4\% and 15.8\% lower than REINFORCE  and DDQN respectively. The main reason behind this is that the A3C-R2N2 network is able to adapt to the task workload behavior quickly. This allows a resource hungry task to be scheduled to a powerful machine. Moreover, the presence of Average Energy Consumption (AEC) metric of all the edge-cloud nodes within the loss function enforces the model to take energy efficient scheduling decisions. It results in the minimum number of active hosts with the remaining hosts in stand-by mode to conserve energy (utilizing this feature of CloudSim). Moreover, Figure \ref{fig:fcn-response} shows that among all the scheduling policies, A3C-R2N2 provides the least average response time which is 7.74\% lower than the REINFORCE policy, best among the baseline algorithms. This is because the A3C-R2N2 model explicitly takes input about whether a node is a edge or cloud node and allocates tasks without multiple migrations and Average Migration Time (AMT) being embedded in the loss function. As shown in Figure \ref{fig:fcn-sla}, the A3C-R2N2 model has the least number of SLA violations which is 31.9\% lower than the REINFORCE policy. This again is due to reduced migrations and intelligent scheduling of tasks to prevent the high loss value because of SLA violations. As shown in Figure \ref{fig:fcn-cost}, the total cost of the data center is least for the A3C-R2N2 model as it gets the cost model (Cost per hour consumption) for each host as a feature in $FV_i^{Hosts}$ and can ensure that tasks can be allocated to as low number of cloud VMs as possible to reduce cost. Compared to the best baseline (REINFORCE), the A3C-R2N2 model reduces cost by 4.64\%. 

Furthermore, the A3C-R2N2 model also considers the tasks completion time in the previous scheduling interval and the expected completion time for running tasks. For time-critical tasks, the A3C-R2N2 model allocates it  to a powerful host machine and avoid migration to save the migration time. This way, the A3C-R2N2 model can reduce the average completion time as shown in Figure \ref{fig:fcn-completion} which is lower than REINFORCE by 17.53\%. Also, as seen in Figure \ref{fig:fcn-tasks}, the number of tasks completed and the fraction completed in expected time is highest for the A3C-R2N2 model. As a number of migration and migration time severely affect the quality of response of the tasks, Figure \ref{fig:fcn-migrations} and \ref{fig:fcn-migration-time} show how A3C-R2N2 model can achieve the best metric values by having a low number of task migrations. 


To compare the scheduling overhead of the R2N2 model with the baseline algorithms, we provide a comparative result in Figure \ref{fig:overhead}. As the R2N2 network needs to be updated every 1 hour of simulation time, the scheduling time is slightly higher than the other algorithms. Heuristic-based algorithms have very low scheduling overhead as they follow simple greedy approaches. R2N2 model has overhead higher by 0.002\% from RL model. Even though the scheduling overhead is higher than the baseline algorithms, it is not significantly large. Considering the performance improvement by  the R2N2 model, this overhead is negligible and makes the R2N2 model  a better scheduler compared to the heuristics or traditional RL based techniques for Edge-Cloud environments with stochastic workloads.

\begin{figure}
\centering
  \includegraphics[width=0.5\linewidth]{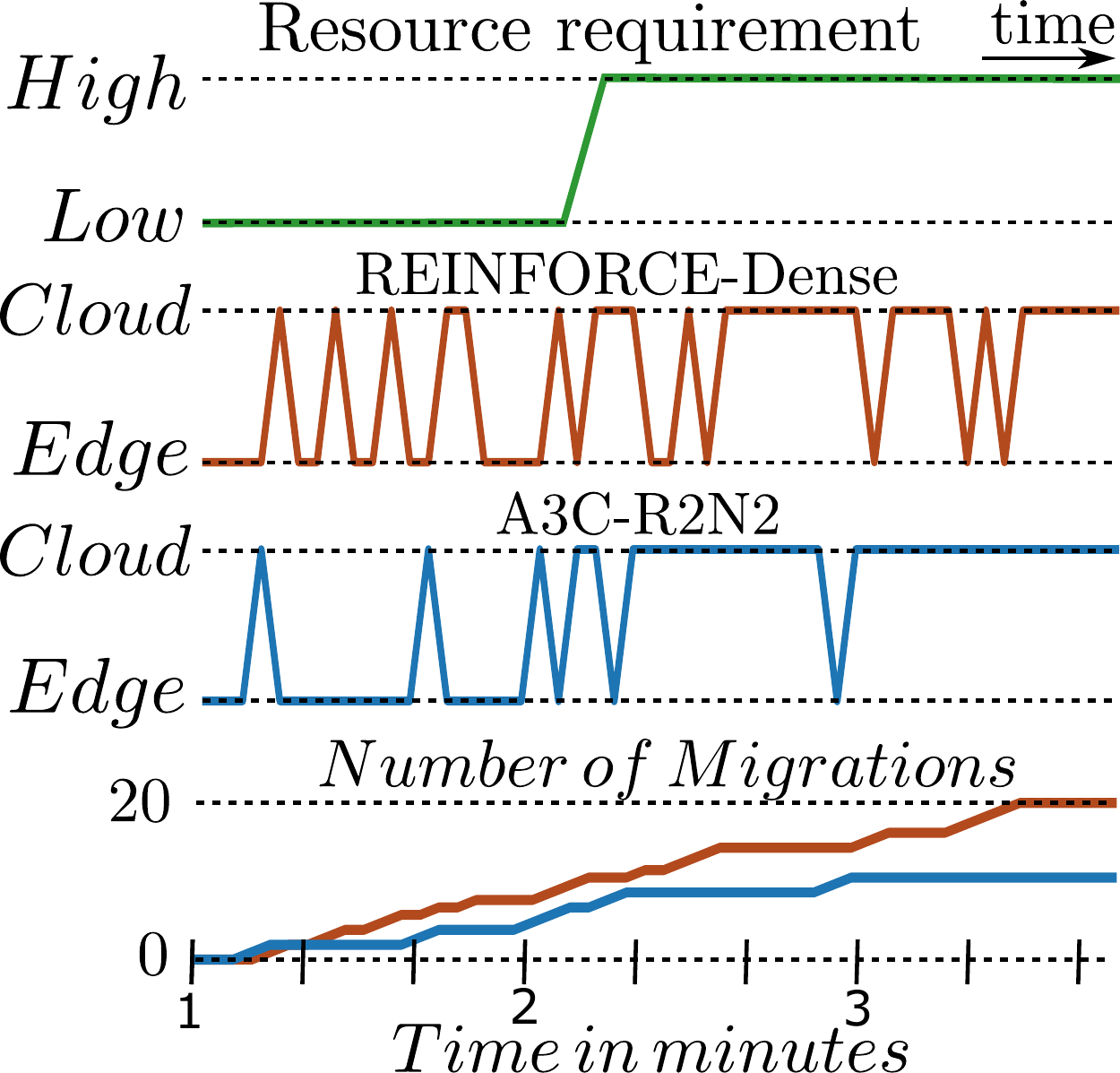}
    \caption{Allocation timeline}
    \label{fig:rnn-comparison}
\end{figure}


\subsection{Summary of insights}

\begin{table*}[!t]
   \centering
    \resizebox{\textwidth}{!}{
\begin{tabular}{|c|c|c|c|c|c|c|c|c|c|c|c|}
\hline 
\multirow{2}{*}{Work} & \multicolumn{1}{c|}{Edge} & \multirow{2}{*}{Decentralized} & \multirow{1}{*}{Hetero-} & \multirow{2}{*}{Dynamic} & Stochastic & Adaptive & \multirow{2}{*}{Method} & \multicolumn{4}{c|}{Optimization Parameters}\tabularnewline
\cline{9-12} 
 & Cloud &  & geneous &  & Workload & QoS &  & Energy & Response Time & SLA Violations & Cost\tabularnewline
\hline 
\hline 

    \cite{beloglazov2012optimal} & \xmark & \xmark  & \cmark & \cmark  & \xmark  & \xmark  & Heuristics  & \cmark & \xmark & \cmark  & \xmark \tabularnewline
\hline 
    \cite{pham2016fog} & \cmark & \xmark  & \cmark & \xmark  & \xmark  & \xmark  & Heuristics  & \cmark & \xmark & \xmark  & \cmark \tabularnewline
\hline 
 
 \cite{bui2017energyJPDC}    & \xmark & \xmark  & \cmark   & \cmark & \cmark & \xmark & Gaussian Process Regression  & \cmark  & \xmark  & \cmark & \xmark \tabularnewline

\hline 

\cite{cheng2018drlcloud,huang2019}    & \xmark & \xmark & \cmark  & \cmark  & \cmark & \cmark & DQN  & \cmark & \xmark  & \xmark  & \cmark \tabularnewline
\hline 
  \cite{basu2019Megh}  & \xmark & \xmark  & \cmark  & \cmark & \cmark & \cmark & Q Learning & \cmark  & \xmark  & \xmark & \cmark \tabularnewline
\hline
\cite{xu2017DLlaserdeadline}    & \xmark & \xmark  & \xmark & \cmark  & \cmark & \xmark & DNN  & \xmark  &\xmark  & \cmark  & \cmark \tabularnewline
\hline 
  \cite{mao2016RMSDRL, zhang2018doubleQLearningEdge,li2019deepjs}  & \xmark & \xmark  & \cmark & \cmark &\cmark  & \cmark  & DDQN  & \cmark &\xmark  & \xmark  & \xmark \tabularnewline
\hline 
  \cite{mao2016resource,rjoubdeep}   & \xmark & \xmark  &  \cmark & \cmark & \cmark  & \cmark  & DRL (REINFORCE)   & \xmark  & \cmark & \xmark & \xmark\tabularnewline
\hline
\textbf{This Work}    & \cmark & \cmark  & \cmark & \cmark  & \cmark & \cmark & DRL (A3C-R2N2)  & \cmark  &\cmark  & \cmark  & \cmark \tabularnewline

\hline 
\end{tabular}
}
\caption{Comparison of Related Works with Different Parameters}
\label{relatedworktable}
\end{table*}

\begin{figure}
    \centering
    \includegraphics[width=0.45\linewidth]{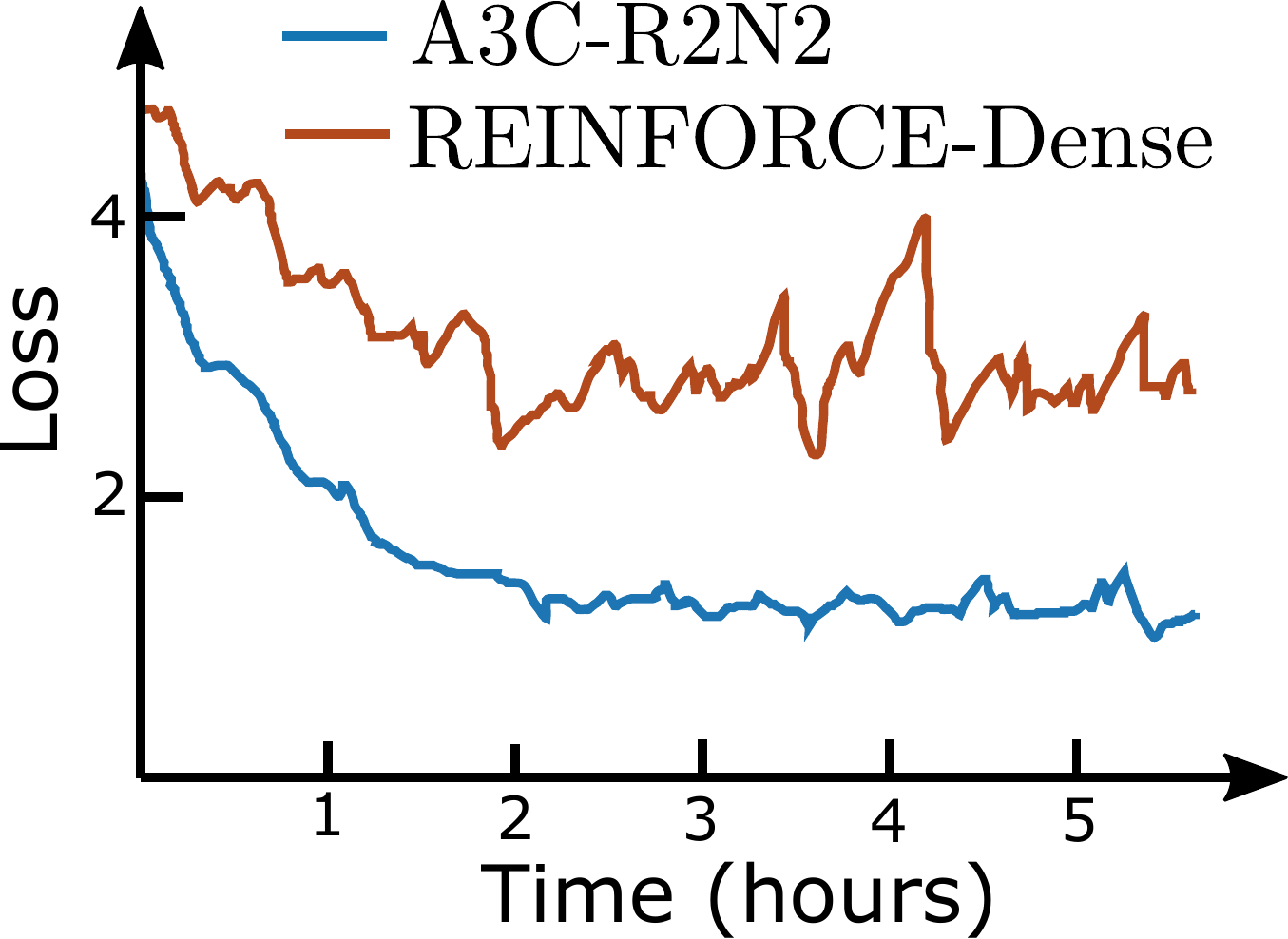}
    \caption{Convergence comparison}
    \label{fig:convergence}
\end{figure}

The R2N2 model works better than the baseline algorithms because it can sense and adapt to the dynamically changing environment, unlike the heuristic-based policies which use a representative technique for making scheduling decisions and are prone to jump to erroneous conclusions due to their limited adaptability. Compared to the DDQN approach, asynchronous policy gradient allows the R2N2 model to quickly change the scheduling policy based on changes in network, workload and device characteristics allowing the model to quickly adapt to dynamically changing scenarios. Figure \ref{fig:rnn-comparison} shows scheduling decisions classified as edge or cloud for different approaches with time for a sample task and response time minimization goal. For a task that has low resource requirement, it is better to schedule in low latency edge node rather than cloud. When task becomes resource intensive, only then is it optimal to send it to cloud as it may slow down the edge node. The REINFORCE-Dense model is unable to exploit temporal patterns like increasing resource utilization of a task with previous scheduling decisions to optimally decide the task allocation. This not only leads to higher frequency of sub-optimal decisions but also increases migration time. Considering these points, the A3C-R2N2 strategy can adapt to non-stationary targets and approximate and learn the parameters much faster and more precisely compared to the traditional RL based approaches as shown in Figure \ref{fig:convergence}. Figure~\ref{fig:convergence} also shows that the loss value for the RL framework is much lower when the A3C-R2N2 model compared to the REINFORCE-Dense model. The average loss value in last 1 hour in a full day experiment is 2.78 for REINFORCE-Dense and 1.12 (nearly 60\% reduction in loss value) for the proposed model. To summarize, earlier works did not model temporal aspects using neural networks due to slower training of recurrent layers like GRU. However, modern advancements of residual connections and the proposed formulation allow faster propagation of gradients leading to a solution for the slow training problem. 

\section{Related Work}
\label{sec:related-work}


Several studies \cite{skarlat2017optimized, beloglazov2012optimal, pham2017cost, brogi2017qos, choudhari2018prioritized, pham2016fog, xiong2019deep} have proposed different types of heuristics for the scheduling applications in Edge-Cloud environment. Each of these studies focuses on optimizing different parameters for a specific set of applications. Some of the works are applied to Cloud systems, while others are for Edge-Cloud environments. It is well known that heuristics work for generic cases and fail to respond to the dynamic changes in environments. However, a learning-based model can adapt and improve over time by tuning its parameters according to new observations.  

Predictive optimizations have been studied by \cite{cheng2018drlcloud, basu2019Megh, mao2016RMSDRL, li2018IoTlearning, zhang2017energydeepQRL,xu2017DLlaserdeadline, zhang2018doubleQLearningEdge, huang2019} in many of the recent works. These works use different ML (Machine Learning) and DL (Deep Learning) techniques to optimize the Resource Management System (RMS). Deep Neural Networks (DNN) and Deep Reinforcement Learning (DRL) approaches have been widely used in this regard.  In most of these works, optimizing energy is a primary objective. Bui et al. \cite{bui2017energyJPDC} studied a predictive optimization framework for energy efficiency of cloud computing. They predict the resource utilization of the system in the next scheduling period by Gaussian process regression method. Based on this prediction, they choose a minimum number of servers to be active to reduce the energy consumption of the overall system. However, their approach still uses many heuristics in scheduling decisions and hence do not adapt to dynamic Edge-Cloud environments or changing workload characteristics.    
Zhang et al, \cite {zhang2017energydeepQRL} proposed a DDQN for energy-efficient edge computing. The proposed hybrid dynamic voltage frequency scaling (DVFS) scheduling based on Q-learning.   As a deep Q-learning model cannot distinguish the continuous system states, in an extended work \cite {zhang2018doubleQLearningEdge}, they investigated a double deep Q-learning model to optimize the solution further.  Xu et al. \cite{xu2017DLlaserdeadline} proposed LASER, a DNN approach for speculative execution and replication of deadline critical jobs in the cloud. They implement these DNN based scheduling framework for the Hadoop framework. Basu et al. \cite{basu2019Megh}  investigated the live migration problem of Virtual Machines (VMs) using RL based Q-learning model. The proposed algorithms are aimed to improve over existing heuristic-based live migration. Live migration is widely used for consolidating the VMs to reduce energy consumption. Their proposed RL model – Megh,  continuously adapts and learns to the changes in the system to increase the energy efficiency.  Cheng et al.\cite{cheng2018drlcloud} have studied Deep reinforcement learning-based resource provisioning and task scheduling approach for cloud service providers. Their Q-learning based model is optimized to reduce the electricity price and task rejection rate. Similarly, Mao et al.\cite{mao2016RMSDRL} and Li et al.\cite{li2019deepjs} explored Resource Management with DDQN. They apply the DRL to scheduling jobs on multiple resources and analyze the reasons for achieving high gain compared to state-of-the-art heuristics. As described before, these Q-learning based algorithms lack the ability to quickly adapt in stochastic environments. Mao et al. \cite{mao2016resource} and Rjoub et al.\cite{rjoubdeep} also explored DRL (REINFORCE) based scheduling for edge only environments. They only consider response time as a metric and also do not exploit asynchronous or recurrent networks to optimize model adaptability and robustness.

A summary of the comparison of relevant works with our work over different parameters is shown in Table \ref{relatedworktable}. We consider that the scheduler is dynamic if the optimization is carried dynamically for active tasks and new tasks that arrive in the system continuously. Stochastic workload is defined by changing tasks arrival rates and resource consumption characteristics. The definitions for remaining  parameters are self explanatory. \black{For the sake of brevity, instead of comparing to all the heuristics based work in the table, we compare our work to \cite{beloglazov2012optimal} and \cite{pham2016fog} which act as some of the baseline algorithm in our experiments. The existing RL based solutions use  Q-learning models \cite{basu2019Megh, cheng2018drlcloud, mao2016RMSDRL} and are focused on optimizing the specific parameters such as energy or cost, wherein we compare our approach with DDQN~\cite{li2019deepjs} and DRL (REINFORCE)~\cite{rjoubdeep}. All these baseline methods are adapted to be used in the proposed edge-cloud setup.} However, in the Edge-Cloud environments, infrastructure is shared among the diverse set of users requiring different QoS for their respective applications. In such a case, the scheduling algorithm must be adaptive and be able to tune automatically to application requirements. Our proposed framework can be optimized to achieve better efficiency with respect to different QoS parameters as shown in Section 4 and Section 5. Moreover, Edge-Cloud environment brings heterogeneous complexity and stochastic behavior of workloads which need to be modeled within a scheduling problem. We model these parameters efficiently in our model.

\section{Conclusions and Future Work}
\label{sec:conclusion}
Efficiently utilizing edge and cloud resources to provide a better QoS and response time in stochastic environments with dynamic workloads is a complex problem. This problem is complicated further due to the heterogeneity of multi-layer resources and difference in response times of devices in  Edge-Cloud datacenters. Integrated usage of cloud and edge is a non-trivial problem as resources and network have completely different characteristics when users or edge-nodes are mobile. Prior work not only fails to consider these differences in edge and cloud devices but also ignores the effect of stochastic workloads and dynamic environments. This work aims to provide an end-to-end real-time task scheduler for integrated edge and cloud computing environments. We propose a novel A3C-R2N2 based scheduler that can consider all important parameters of tasks and hosts to make scheduling decisions to provide better performance. Furthermore, A3C allows the scheduler to quickly adapt to dynamically changing environments using asynchronous updates, and R2N2 is able to quickly learn network weights also exploiting the temporal task/workload behaviours. Extensive simulation experiments using iFogSim and CloudSim on real-world Bitbrain dataset show that our approach can reduce energy consumption by 14.4\%, response time by 7.74\%, SLA violations by 31.9\% and cost by 4.64\%. Moreover, our model has a negligible scheduling overhead of  0.002\% compared to the existing baseline which makes it a better alternative for dynamic task scheduling in stochastic environments. 

As part of future work, we plan to implement this model in real edge-cloud environments. Implementation in real environments would require constant profiling CPU, RAM and disk requirements of new tasks. This can be done using exponential averaging of requirement values in the current scheduling interval with the average computed in the previous interval. Further, the CPU, RAM, disk and bandwidth usage would have to be collected and synchronized across all A3C agents in the edge-cloud setup. Further to the scalablity analysis, we also plan to conduct tests to check the scalability of the proposed framework with number of hosts and tasks. The current model can schedule for a fixed number of edge nodes and tasks. However, upcoming scalable reinforcement learning models like Impala \cite{espeholt2018impala} can be investigated in future. Moreover, we plan to investigate the data privacy and security aspects in future.

\section*{Software Availability}
Our code, experiment scripts and raw result files are available online under GPL-3.0 License at: \url{https://github.com/Cloudslab/DLSF}.

\section*{Acknowledgements}
This research work is supported by the Melbourne-Chindia Cloud Computing (MC3) Research Network and the Australian Research Council.

\bibliographystyle{IEEEtran}

\bibliography{references}

\vspace{-0.5in}

\begin{IEEEbiography}
[{\includegraphics[width=1in,height=1in,clip,keepaspectratio]{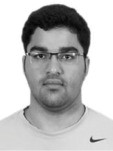}}]
{Shreshth Tuli}
is an undergraduate student at the Department of Computer Science and Engineering at Indian Institute of Technology - Delhi, India. He worked as a visiting research fellow at the CLOUDS Laboratory, School of Computing and Information Systems, the University of Melbourne, Australia. His research interests include Internet of Things (IoT), Fog Computing, Blockchain, and Deep Learning.
\end{IEEEbiography}
\vspace{-0.5in}
\begin{IEEEbiography}
[{\includegraphics[width=1in,height=1in,clip,keepaspectratio]{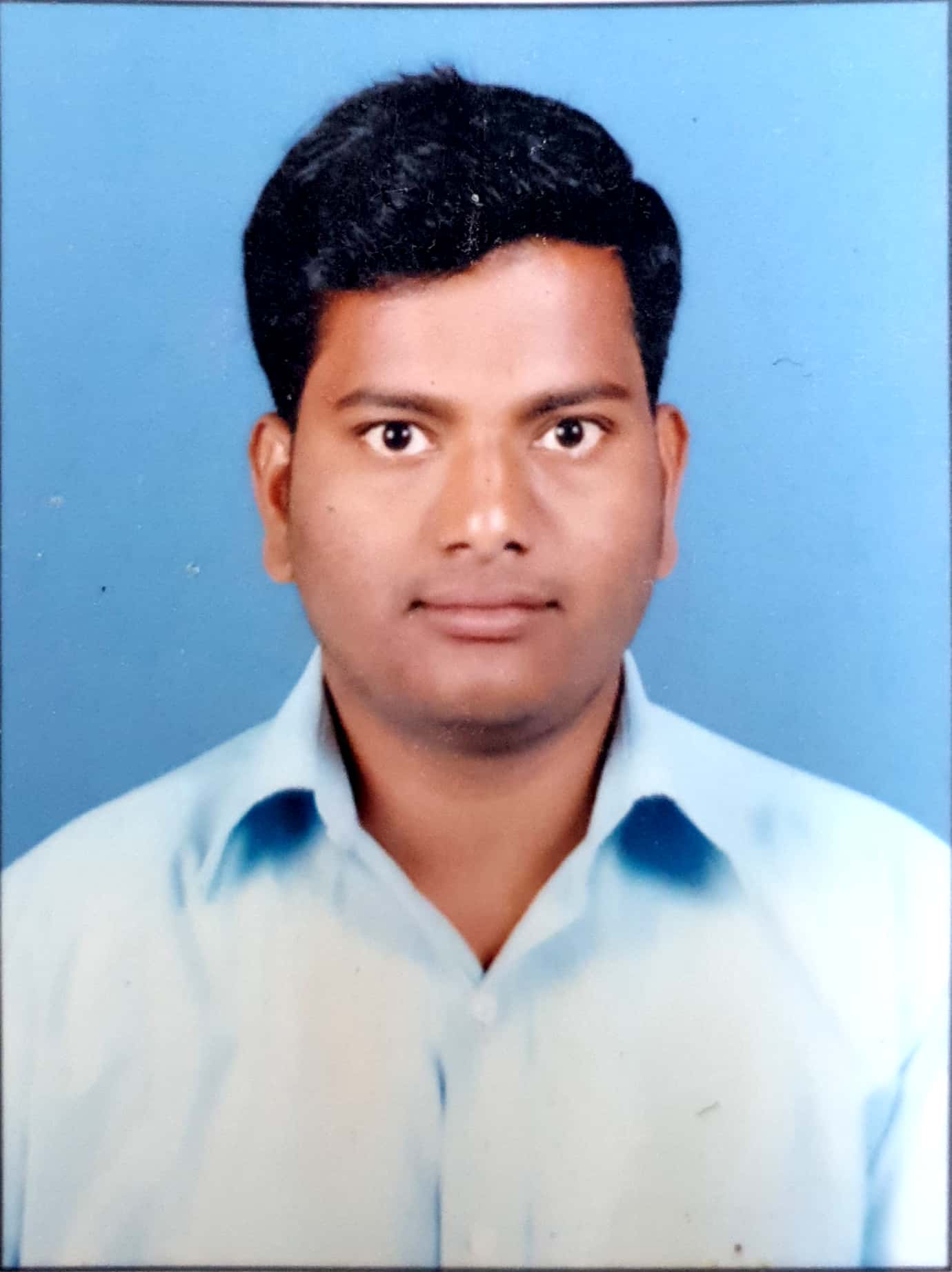}}]
{Shashikant Ilager}
is a PhD candidate with the CLOUDS Laboratory at the University of Melbourne, Australia. His research interests include distributed systems and cloud computing. He is currently working on resource management through data-driven predictive optimization techniques in large scale distributed systems.
\end{IEEEbiography}
\vspace{-0.5in}
\begin{IEEEbiography}
 [{\includegraphics[width=1in,height=1in,clip,keepaspectratio]{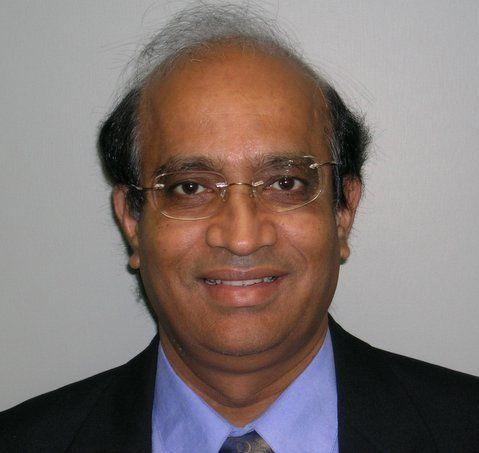}}]
    {Kotagiri Ramamohanarao }
received the PhD degree from Monash University. He is currently a professor of computer science with the University of Melbourne. He served on the editorial boards of the Computer Journal. At present, he is on the editorial boards of Universal Computer Science, Data Mining, and the International Very  Large Data Bases Journal. He was the program co-chair for VLDB and DASFAA conferences.
\end{IEEEbiography}
\vspace{-0.5in} 
\begin{IEEEbiography}
  [{\includegraphics[width=1in,height=1in,clip,keepaspectratio]{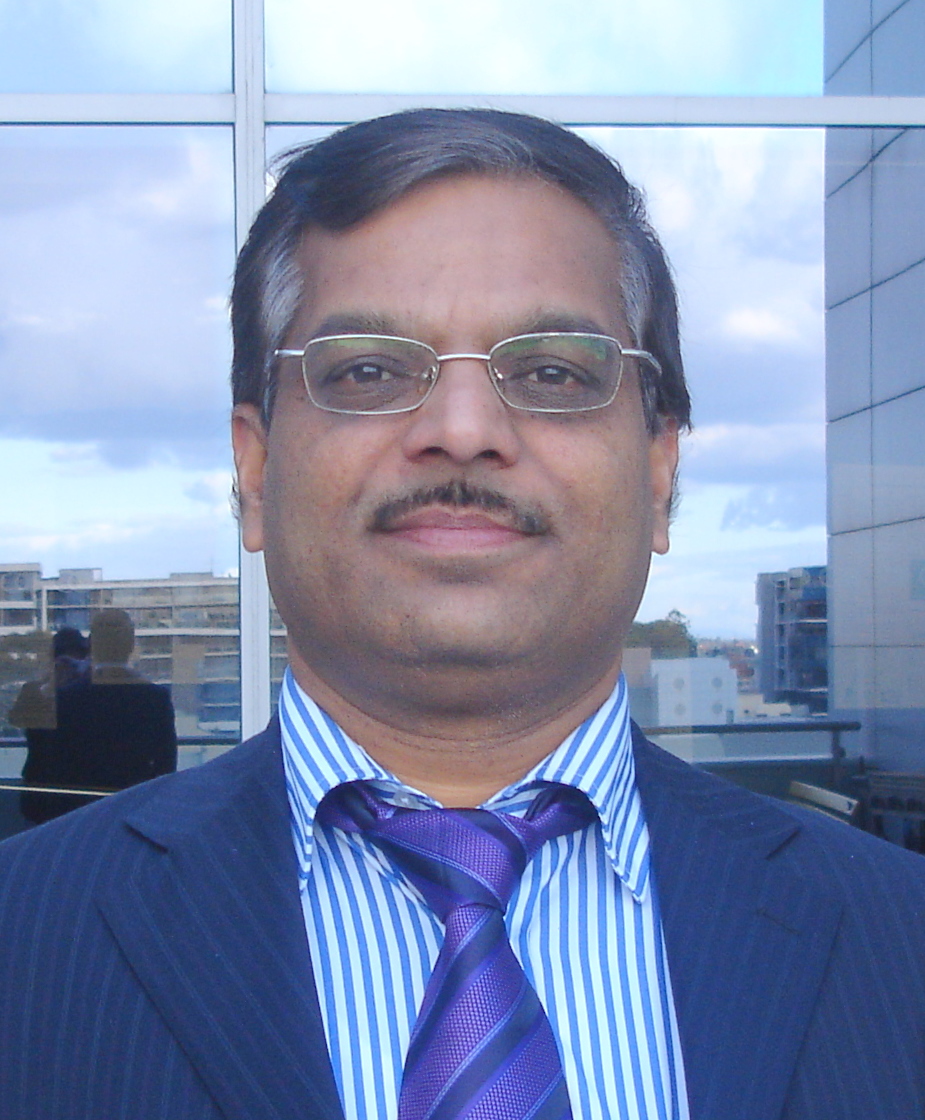}}]
    {Rajkumar Buyya} is a Redmond Barry Distinguished Professor and Director of the Cloud Computing and Distributed Systems (CLOUDS) Laboratory at the University of Melbourne, Australia. He has authored over 725 publications and seven textbooks including "Mastering Cloud Computing" published by McGraw Hill, China Machine Press, and Morgan Kaufmann for Indian, Chinese and international markets respectively. He is one of the highly cited authors in computer science and software engineering worldwide (h-index=137, g-index=304, 99,800+ citations).  He is a fellow of the IEEE.
\end{IEEEbiography}
\vfill

\end{document}